\documentclass[nonacm,sigplan,10pt]{acmart}
\usepackage{microtype}
\usepackage{graphicx}
\usepackage{booktabs} 
\usepackage{tabularx}

\usepackage{multirow}
\usepackage{makecell}
\usepackage{hhline}
\usepackage{xcolor}
\usepackage{bm}
\usepackage{tikz}
\usepackage{amsmath}
\usepackage{enumitem}
\usepackage{caption}
\usepackage[ruled,linesnumbered,noend]{algorithm2e}

\SetCommentSty{mycommfont}
\usepackage{amsthm}
\usepackage{ifthen}
\usepackage{comment}

\usepackage{listings}
\definecolor{deepblue}{rgb}{0,0,0.5}
\definecolor{deepred}{rgb}{0.6,0,0}
\definecolor{deepgreen}{rgb}{0,0.5,0}
\usepackage{url}
\usepackage{subcaption}
\newcommand{\bluecomment}[1]{\textcolor{blue}{\textit{// #1}}}

\usepackage{hyperref}
\newcommand{\sys}{BROS}

\begin{document}

\title{Efficient LLM Serving on Hybrid Real-time and Best-effort Requests}

\author{Borui Wan*}
\affiliation{
  \institution{The University of Hong Kong}
  \country{}
}
\email{wanborui@connect.hku.hk}

\author{Juntao Zhao*}
\affiliation{
  \institution{The University of Hong Kong}
  \country{}
}
\email{juntaozh@connect.hku.hk}

\author{Chenyu Jiang}
\affiliation{
  \institution{The University of Hong Kong}
  \country{}
}
\email{jchenyu@connect.hku.hk}

\author{Chuanxiong Guo}
\affiliation{
  \institution{BitIntelligence}
  \country{}
}
\email{chuanxiong.guo@gmail.com}

\author{Chuan Wu}
\affiliation{
  \institution{The University of Hong Kong}
  \country{}
}
\email{cwu@cs.hku.hk}

\thanks{* Equal contribution}
\begin{abstract}
Recent breakthroughs in large Language Models (LLMs)
have enabled various generative tasks on a single model. Real-world services (e.g., OpenAI's ChatGPT~\cite{chatgpt}) powered by an LLM often concurrently support latency-critical requests for interactive applications (e.g., question-answering systems, referred to as \textit{real-time} or \textit{RT} requests) and throughput-oriented requests for back-of-house processing (e.g., documents batch processing~\cite{batch-api}, referred to \textit{best-effort} or \textit{BE} requests), with complex hybrid inference workloads to the underlying model.
State-of-the-art (SOTA) LLM serving systems dedicate machines to each type of request, towards either low inference latency or high serving throughput, respectively.
This practice simplifies request scheduling and management but suffers from poor resource utilization.
We propose \sys{}, a hybrid LLM serving system that aims to collocate RT/BE requests, meeting RT requests' latency requirements while maintaining BE requests' throughput.
\sys{} 
formulates the problem of hybrid RT/BE request scheduling and solves it with a dynamic priority-based algorithm.
\sys{} designs a bidirectional KV cache management mechanism, allowing RT requests to share KV memory with BE requests to remove the scheduling restrictions caused by insufficient KV memory and improve utilization.
Extensive experiments validate that \sys{} achieves a good trade-off when serving hybrid RT and BE requests. It significantly reduces the latency of RT requests (up to 74.20\%), improving their fine-grained service level objectives (SLOs) attainments (up to 36.38$\times$), with negligible throughput reduction for BE requests, showing significant advantages over SOTA systems like vLLM and TGI.
\end{abstract}

\maketitle

\section{Introduction}
\label{sec:intro}

\begin{figure}[t]
  \centering
  \includegraphics[width=\linewidth]{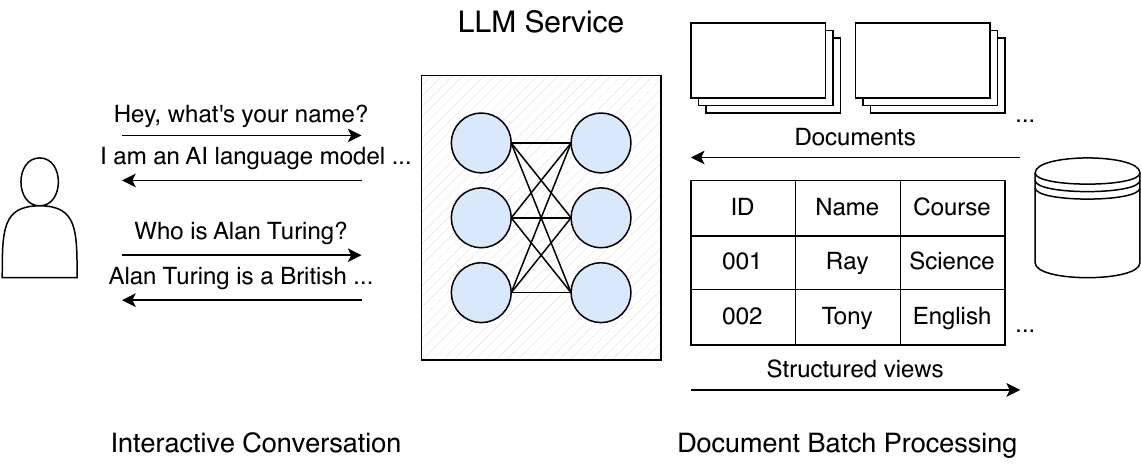}
  \caption{
A powerful LLM can serve real-time interactive services while also handling vast non-real-time back-of-house tasks.
}
\label{fig:tasks}
\end{figure}

Large Language Models (LLMs) such as GPT~\cite{gpt}, Gemini~\cite{gemini}, Llama~\cite{llama, llama3} and DeepSeek~\cite{deepseek-v3}, have demonstrated remarkable performance on language-related generative tasks, enabling various real-world online applications such as role-playing~\cite{character-ai}, creative helper~\cite{recipes, chatgpt}, programming assistants~\cite{copilot}, and offline applications including document batch processing~\cite{batch-api}, information extraction~\cite{datalakes}, and data wrangling~\cite{datawrangling}. A single state-of-the-art (SOTA) LLM can handle various tasks wisely with proper supervised fine-tuning~\cite{wei2021sft, wang2022sft}, reinforcement learning~\cite{ppo, grpo}, and prompting engineering~\cite{prompt_survey}.
For example, the text-davinci series LLMs~\cite{sharegpt}, accessed via OpenAI's API, can serve as an online chatbot as well as a document processor (As depicted in Fig.~\ref{fig:tasks}).

Online applications such as ChatGPT~\cite{chatgpt} and AI anime chatbots~\cite{character-ai} return the generated context in a streaming manner~\cite{streaming}, demanding both quick response after queueing and high context rate (e.g., 250 words/min)~\cite{distserve} afterward.
Apart from normalized latency\footnote{Normalized latency is the mean of every request's end-to-end latency divided by its output token number.}~\cite{orca, vllm}, this serving goal is also measured by two metrics: time to first token (TTFT) and time per output token (TPOT)~\cite{distserve}.
We refer to such LLM workload as \textit{real-time} or RT requests.
Offline applications, such as the asynchronous document processing service provided by OpenAI's Batch API~\cite{batch-api}, usually involve inference workloads processed in a batch manner within a specified time window (e.g., 24 hours in OpenAI's Batch API).
Therefore, they are not sensitive to latency but favor high throughput (request per second).
We refer to such requests as \textit{best-effort} or BE requests.
This discrepancy in serving goals for RT/BE requests leads to the disaggregated deployments of online and offline LLM services on separated GPU clusters, which is the current common practice for production~\cite{flexgen,distserve,taming, blendserve, llumnix, dynamollm}.  

However, the volume of real-world RT requests commonly changes in a diurnal pattern~\cite{alpa-serve, burstgpt}.
To fulfill the serving goals of RT requests, service providers should provide enough GPU resources to meet the peak demands.
Over-provisioning separate resources results in GPU cycles and energy wastes, both of which finally translate into operational costs.
Co-serving RT and BE requests on shared resources offers a compelling alternative.
By strategically batching latency-sensitive BE requests with RT workloads during opportune intervals, systems can enhance resource utilization while maintaining strict RT latency guarantees.
Nevertheless, LLM's unique autoregressive inference pattern presents unprecedented challenges for hybrid LLM serving.

{\em First}, the non-predetermined number of LLM inference passes makes it difficult to schedule RT/BE requests to meet their distinct serving goals.
Given the quick response and high context rate requirements of RT requests, the desirable property of the LLM serving system is to balance TTFT and TPOT latency Service-Level Objectives (SLOs) of RT requests while maintaining long-term performance for BE requests.
However, existing request scheduling mechanisms fall short of achieving a good trade-off between RT and BE requests.
An LLM responds to user requests with one \textit{prefill} pass and several \textit{decode} passes, where the number of the latter cannot be known at the queuing stage.
Multiple model querying iterations prevent conventional priority fair scheduling from directly assigning priority values (e.g., total inference time or remaining time) for each LLM request.
Other heuristics like Multi-level Feedback Queue (MLFQ), when applied to LLM request scheduling~\cite{fastserve}, can reduce the average end-to-end inference latency.
Nevertheless, this approach does not align well with the quick response demand of RT requests and is not suitable for hybrid requests where BE requests are insensitive to latency.
Moreover, BE and RT requests exhibit conflicting preferences for system configurations, particularly in batch sizing.
RT requests favor smaller batches to minimize latency~\cite{dvabatch, tensorflow-serving, nvidia-triton}, while BE requests benefit from larger batches to maximize GPU utilization and throughput~\cite{orca, vllm}.
These opposing requirements create significant challenges in configuring dynamic RT/BE co-serving systems.

{\em Further}, co-scheduling RT/BE requests brings GPU memory contention associated with caching intermediate states.
As the core of LLMs, decoder-based Transformers~\cite{attention} generate one token in each pass, based on the input prompt tokens and all the previously generated tokens.
The key and value tensors of all previously processed tokens need to be cached in GPU memory for the generation of the next token (aka KV cache).
Concurrent RT and BE requests compete for memory resources to store their KV cache, which imposes limitations on scheduling decisions.
As the \textit{first-class citizen}, high-priority RT requests may fail to be scheduled when available GPU memory is insufficient to accommodate their increasing KV cache.
Simply dropping or swapping the KV cache~\cite{vllm, fastserve} of low-priority BE requests leads to recomputation or Device-to-Host (D2H) copy overheads, delaying their completion of inference. 

None of the existing LLM serving systems~\cite{orca, vllm, flexgen, fastserve, llumnix, distserve} designs dedicated hybrid RT/BE requests scheduling mechanisms.
Most of them~\cite{orca, TGI, llumnix, distserve} still rely on First-Come-First-Serve (FCFS) scheduling, which can easily incur head-of-line blocking due to running early-arrived BE requests with long prompts and output sequences that block late-joining RT requests.
Besides, the KV cache management strategies of existing systems do not handle the storage competition between RT and BE requests well.

We propose \sys{}, an LLM serving system tailored for handling hybrid RT and BE requests. \sys{} advocates fine-grained request scheduling at the iteration level to ensure two latency SLOs for RT requests and maximize inference throughput for BE requests, as well as effectively manage the available memory resources between RT and BE requests. Our contributions are summarized as follows:

$\triangleright$ 
We formulate the hybrid scheduling problem and propose a dynamic packing algorithm for hybrid request scheduling.
To fulfill quick response and high context rate demands of RT requests, the hybrid scheduling problem views each queueing RT request's TTFT and TPOT metrics as the constraints while optimizing the throughput of BE requests to avoid starvation. 
The proposed heuristic first batches RT requests according to their dynamically assigned priorities and then adaptively replaces low-priority RT requests with BE requests.
We further provide an SLO-aware adaptive batch sizing mechanism to control the runtime batch size.
As a result, \sys{} strikes a favorable serving trade-off between RT and BE requests.

$\triangleright$
We resolve memory competition between RT and BE requests with a bidirectional storage layout design for the KV cache. The available memory is divided into blocks~\cite{vllm}.
Each block can be shared by two request types and KV caches of the RT and BE requests expand in opposite directions in each block.
Based on this layout, we propose block preemption to guarantee sufficient GPU memory for high-priority RT requests, along with a lazy checkpointing technique to avoid/delay runtime D2H requirements.

$\triangleright$ We evaluate \sys{} on different workloads with mainstream LLMs 
and datasets. 
Experimental results show that \sys{} significantly reduces the latency of RT requests by 74.20\% maximum, outperforming SOTA production-grade LLM serving systems like vLLM~\cite{vllm} and Text-Generation-Inference (TGI)~\cite{TGI}.

\section{Background and Motivation}\label{sec:bg}
In this section, we first briefly introduce the characteristics of LLM inference (Sec.~\ref{sec:llm_pattern}), then analyze critical limitations of existing serving systems.
These observations motivate \sys{}'s design rationale (Sec.~\ref{sec:llm_chan}).

\subsection{LLM Inference}\label{sec:llm_pattern}
\begin{figure}[!t]
  \centering
  \includegraphics[width=\linewidth]{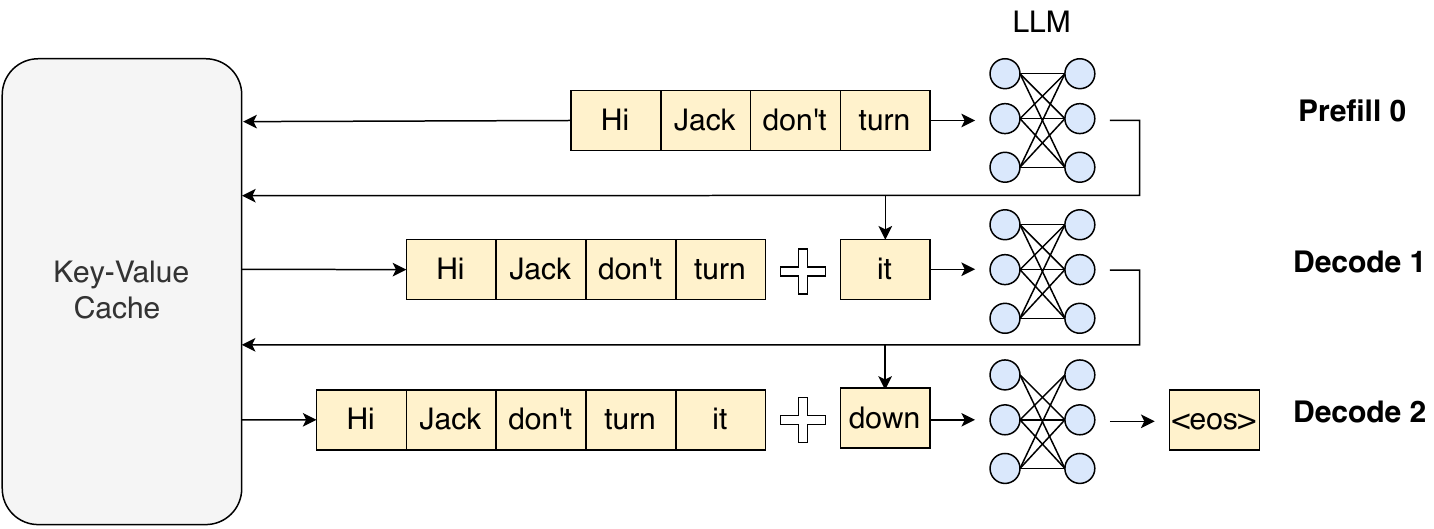}
  \caption{The two phases of an LLM inference request. 
  }
  \label{fig:inf}
\end{figure}
Today's LLMs are empowered by decoder-based Transformer structures~\cite{gpt}. Inference over an LLM is conducted in the 
autoregressive manner, including two phases: prefill and decode. One forward pass over the LLM model is called one \textit{iteration}. As depicted in Fig.~\ref{fig:inf}, the prefill phase involves a single iteration where the prompt tokens (e.g., "Hi Jack, don't turn") are used to compute the first output token ("it"), and all the key-value tensors of the tokens are stored as the initial KV cache. At each iteration of the decode phase, the key-value tensors of the prompt and previously generated tokens are retrieved from the KV cache, and 
utilized, along with 
the latest token, to generate the new token ("down"). If there are more tokens to generate, 
the key-value tensors of the latest token ("it") are also cached. This process continues until the "EOS" token is generated or a maximum output length is reached. Consequently, the total serving time of each request is determined by the prompt length and the number of iterations in the decode phase, which varies from request to request.  

\subsection{Challenges and Opportunities}\label{sec:llm_chan}

\begin{figure}[!t]
    \centering
     \begin{subfigure}[b]{0.49\linewidth}
     \centering
     \includegraphics[width=\linewidth]{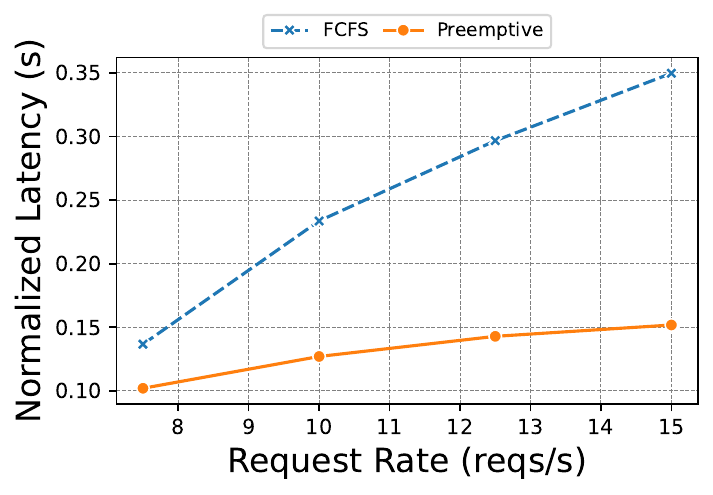}
     \caption{Normalized latency curves.}
     \label{fig:normalized_latency_13B}
    \end{subfigure}
     \hfill
     \begin{subfigure}[b]{0.49\linewidth}
         \centering
         \includegraphics[width=\linewidth]{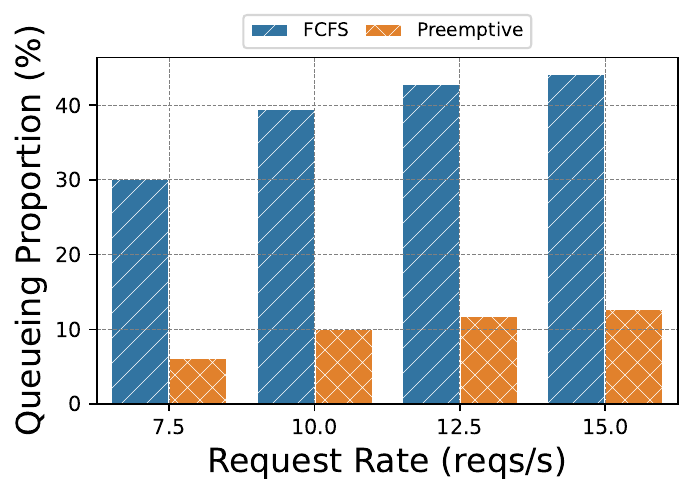}
         \caption{Queueing proportion.}
         \label{fig:queue_time_13B}
     \end{subfigure}
     \caption{
     Head-of-line blocking with FCFS and the benefit of preemptive scheduling.
     The queueing proportion is the mean of every request's queueing time divided by its end-to-end latency.
  }
      \label{fig:head_of_line}
\end{figure}

\noindent\textbf{Observation \#1: Naive preemption benefits RT requests but fails to balance TTFT and TPOT.} 
Most of the existing LLM serving systems~\cite{orca, vllm, distserve} employ FCFS policy to schedule requests, which results in head-of-line blocking when BE requests containing a significant number of output tokens are scheduled ahead of the RT requests.
One possible solution to alleviate this is to allow request preemption at fine granularity, i.e., every LLM forward pass (iteration).
Fig.~\ref{fig:head_of_line} shows the benefits of applying preemptive scheduling with a simple policy, where each request's waiting time between completing one iteration and scheduling the next is used as the priority value.
We run vLLM with FCFS policy and the above preemptive scheduling policy to serve an OPT-13B model on a server with four A100-40GB GPUs, with the tensor-parallel degree of 2.
(detailed configurations in Sec.~\ref{sec:exp}).
512 requests are synthesized based on the SharedGPT~\cite{sharegpt} dataset, then submitted to the system from Poisson distribution with different request rates. 
Compared to the FCFS policy, preemptive scheduling effectively reduces normalized latency (Fig.~\ref{fig:normalized_latency_13B}) by minimizing the overall queuing time (Fig.~\ref{fig:queue_time_13B}) for waiting requests in the system.

\noindent \textbf{Challenges.}
However, existing iterative preemption approaches face a scheduling dilemma to balance TTFT and TPOT SLOs for RT requests.
For example, vLLM preempts decode RT requests for incoming prefill requests, which sacrifices the high context rate (TPOT).
Sarathi-Serve adopts \textit{chunked-prefills} to create stall-free schedules for ongoing decode requests but delays the return of the initial context (TTFT). 

\begin{figure}[!t]
  \centering
  \includegraphics[width=\linewidth]{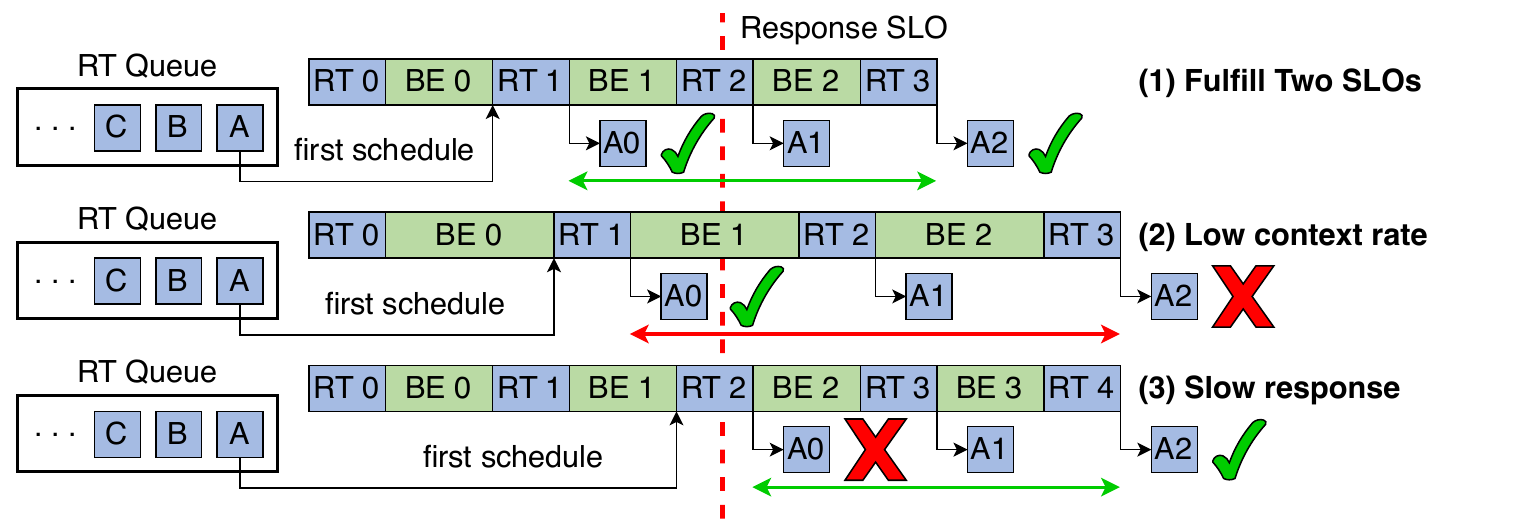}
  \caption{Round-robin scheduling for hybrid RT/BE requests.
  }
  \label{fig:static}
\end{figure}

\noindent\textbf{Observation \#2: Rigid hybrid scheduling is inefficient for co-serving.}
One conventional fair scheduling approach, {\em Iteration-level round-robin scheduling}, is intuitive for co-serving hybrid RT/BE requests.
The system maintains two FCFS queues for RT and BE requests, respectively, and serves them in an interleaving manner at the iteration level.

The round-robin policy allows fair resource sharing between RT and BE requests.
However, this static scheduling mechanism struggles to adapt to LLM requests with varying execution times due to different prompt and output lengths, making it difficult to achieve a favorable serving trade-off. 
We show different cases in Fig.~\ref{fig:static}, where the enqueued RT request A needs three iterations to finish.
The latency SLO for its response (in this example, the time to first token~\cite{distserve}) is denoted by the red dashed line.
The time to generate the subsequent context is represented by double-headed arrows.

In case (1), round-robin scheduling succeeds in providing the quick response and high context rate for request A, while at the same time does not starve BE requests.
However, in case (2), request A's context rate is low due to the prolonged execution time of the BE requests, which may result from more BE requests being scheduled or more tokens per BE request.
In case (3), request A fails to meet the response SLO target since previously scheduled RT requests need more iterations to exit the current batch~\cite{orca}, delaying its first scheduled time. 
Static hybrid scheduling mechanism is inefficient in identifying the most urgent RT requests to serve and balancing RT/BE request serving, according to dynamic request arrivals and their output lengths. 

\noindent \textbf{Challenges.}
Achieving dynamic scheduling is non-trivial: First, due to the varying phases and number of tokens to be processed for different requests, the execution time of each iteration can vary significantly. This variability makes it challenging to determine whether the scheduling decisions may violate the SLO targets. Besides, with a large number of RT and BE requests present in the system, it becomes crucial to select appropriate requests from those combinations for prompt execution.
Failing to do so can result in unacceptable scheduling overhead.
Furthermore, configuring batch sizes to simultaneously meet the divergent requirements of RT and BE workloads is also challenging.

\begin{figure}[!t]
  \centering
  \includegraphics[width=\linewidth]{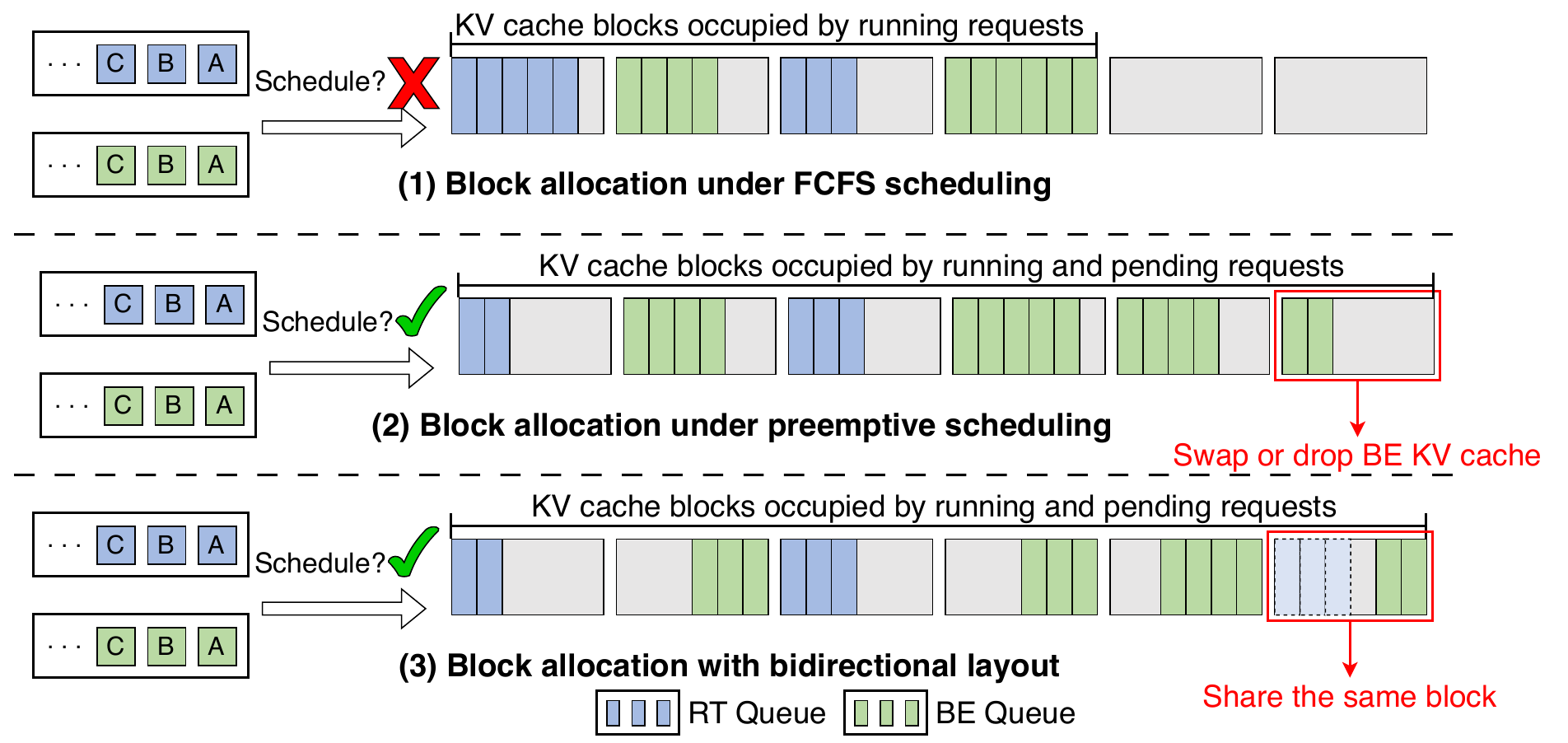}
  \caption{GPU memory block allocation for storing KV cache under different cases.
  }
  \label{fig:bidirectional_motivation}
\end{figure}

\noindent\textbf{Observation \#3: Bidirectional block layout mitigates memory contention.}
Throughout the lifespan of a request in an LLM serving system, the memory demand of its KV cache increases in proportion to the number of iterations it undergoes.
vLLM~\cite{vllm} proposes to dynamically allocate available GPU memory to requests for KV cache storage at the granularity of the block (each contains several slots for tokens' KV tensors), which proves to be an efficient method to improve serving throughput.
With the FCFS schedule, new requests are only served once completed requests exit and release their occupied cache blocks (case (1) in Fig.~\ref{fig:bidirectional_motivation}), regardless of whether they are latency-critical or not;
This results in block wastage when running requests cannot consume all reserved blocks.
When advocating iteration-level request preemption and scheduling among RT and BE requests, there are more concurrent requests in the serving system, 
incurring more KV cache storage overhead and memory competition.
To preemptively schedule later-join RT requests, the system needs to yield cache blocks from BE requests to these RT requests, introducing recomputation or swapping overheads (case (2) in Fig.~\ref{fig:bidirectional_motivation}). 

\noindent \textbf{Opportunities.}
More fine-grained KV cache management assists in reducing memory overheads.
As shown in case (3), a bidirectional block layout, which shares one block between two requests and increases their KV cache in the opposite direction, creates the opportunity for scheduled RT requests to utilize empty slots in the cache block from pending BE requests without KV cache dropping or swapping.

\paragraph{\textbf{Our contributions.}}
In \sys{}, we consider quick response and high context rate requirements for RT requests simultaneously, using them to guide scheduling design for balancing SLO attainments.
We build latency cost models and design an effective scheduling mechanism to batch and run RT and BE requests strategically across iterations.
We properly manage KV cache contention and strategically handle the potential data overwrite issues.

\section{System Overview}\label{sec:design}
\begin{figure}[!t]
  \centering
  \includegraphics[width=0.8\linewidth]{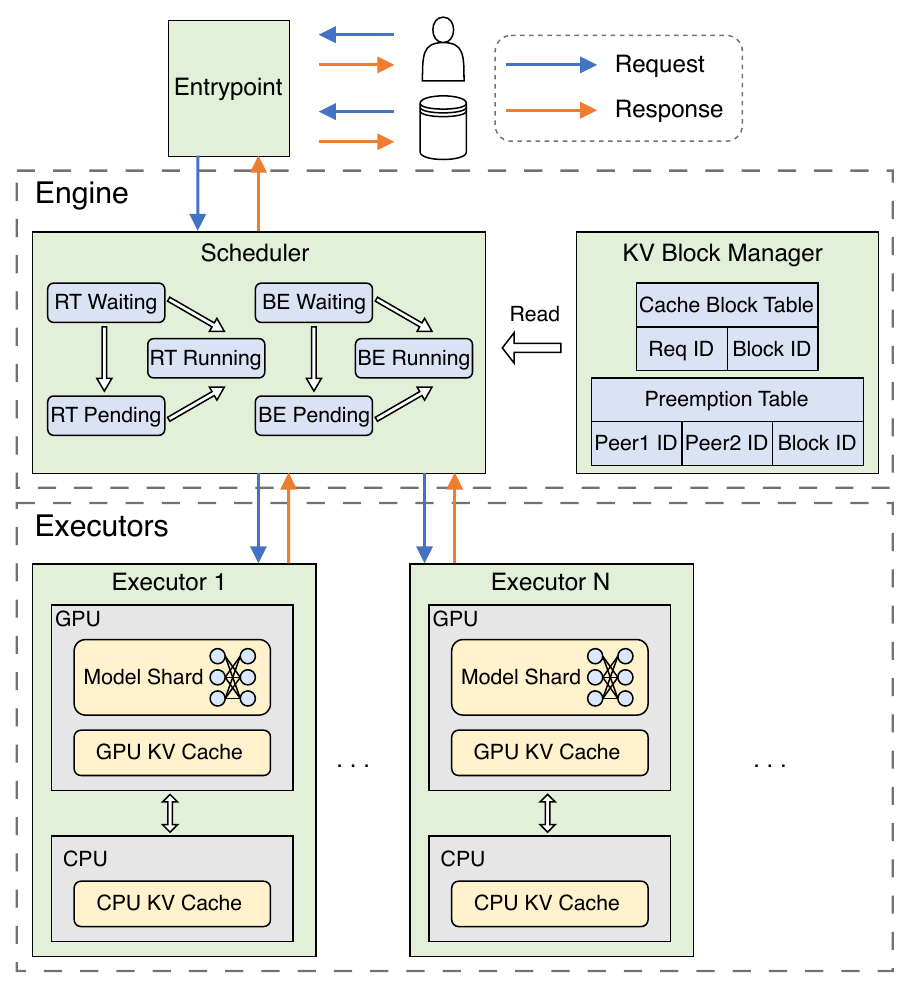}
  \caption{\sys{} system overview.
  }
  \label{fig:arch}
\end{figure}

Fig.~\ref{fig:arch} shows \sys{}'s architecture. It exposes an entrypoint that receives RT (e.g., users' questions) and BE requests (e.g., documents from storage) and forwards them to the scheduling \textit{engine}. Each incoming RT or BE request is initially stored in the corresponding waiting queue and will be moved into the pending queue after its prefill stage processing has been scheduled.
The scheduler running on the engine schedules waiting and pending RT and BE requests at each iteration to populate the respective running queue. 
The \textit{KV block manager} maintains a cache block table that records the mapping between each request ID and the allocated cache block ID. A preemption table is used to track the preemption status of cache blocks. It records the IDs of RT and BE requests that share the same blocks and the IDs of the preempted blocks, indicating occurrence of slot preemption in those blocks. The scheduler acquires KV cache memory utilization and selects appropriate RT and BE requests for execution. 
After scheduling, the engine dispatches the batched requests to the \textit{executors}, where tensor parallelism is supported for distributed inference.
Each executor wraps a GPU and launches a process that loads the weights of the LLM model shard for distributed model execution and manages KV caches of all ongoing requests in GPU and CPU memory. Asynchronous memory-swapping operations are triggered in the executors upon receiving corresponding instructions from the scheduler, to partially checkpoint KV caches of certain requests in CPU memory and swap KV caches of other requests into GPU memory.

\section{Request Scheduler}\label{sec:schedule}
The goal of the scheduler is to define a proper batch size and select suitable RT and BE requests from the waiting and pending queues to execute at each iteration, aiming to meet the latency SLOs of each RT request while maximizing the throughput of BE requests.
In this section, we begin by 
formulate the scheduling problem (Sec.~\ref{sec:problem}), then propose our heuristics and introduce the adaptive batch sizing mechanism in Sec.~\ref{sec:algorithm}.

\subsection{Hybrid Scheduling Problem}\label{sec:problem}

\begin{table}[!t]
\caption{Notation.}\label{tab:notation}
 \resizebox{\linewidth}{!}{
    \begin{tabular}{c|l}
        \toprule
         $x_i$ & whether select RT request $i$ \\
         $y_j$ & whether select BE request $j$ \\
         $t_i$ & elapsed time of RT request $i$ since the last token is returned\\
         $k_i$ & remaining number of tokens in the current token of RT request $i$ \\
         $N^{rt} (N^{be})$ & number of RT (BE) requests in the waiting and pending queues \\
         $L_i^n (L_j^n)$ & input length of RT request $i$ (BE request $j$) \\
         $L_i^a (L_j^a)$ & context length of RT request $i$ (BE request $j$) \\
         $M^{new}_i (M^{new}_j)$ & number of needed new cache blocks of RT request $i$ (BE request $j$)\\
         $M_i^{cpu} (M_j^{cpu})$ & number of checkpointed cache blocks of RT request $i$ (BE request $j$)\\
         $B_{curr}$ & batch size of current iteration\\
         $\mathcal{T}_{SLO}$ & token latency SLO target, which can be either TTFT or TPOT\\
         $T_{avg}$ & the 
         moving-average iteration latency \\
         $M^{ept}$ & number of empty cache blocks \\
         \bottomrule
    \end{tabular}}
    \vspace{-5mm}
\end{table}

\noindent\textbf{Formulation.} We model an iteration-level request batching and scheduling problem, to decide which requests to be included in each batch.
We use decision variables $x_{i}$ to indicate whether the generation of the next token of an RT request $i$ (in the current pending and waiting queues) is included in the current batch, 
and $y_{j}$ to denote whether that of a BE request $j$ is in the batch. 
The goal is to maximize the inference throughput of BE requests 
while ensuring the token latency SLOs of each RT request (notation is summarized in Table~\ref{tab:notation}):  
\begin{equation}\label{eq:obj}
\small
    \max_{x_{i}, y_{j} \in \{0,1\}} \frac{\sum_{j=1}^{N^{be}} y_{j}}{T}
\end{equation}
subject to:
\vspace{-3mm}
\begin{equation}\label{eq:T_total}
\small
    T = C(L^n, L^a, M^{cpu})
\end{equation}
\begin{equation}\label{eq:l^n}
\small
L^n = \sum_{i=1}^{N^{rt}}L_i^n x_{i} + \sum_{j=1}^{N^{be}}L_j^n y_{j}
\end{equation}
\begin{equation}\label{eq:l^a}
\small
L^a = \sum_{i=1}^{N^{rt}}L_i^a x_{i} + \sum_{j=1}^{N^{be}}L_j^a y_{j}
\end{equation}
\begin{equation}\label{eq:M^cpu}
\small
M^{cpu} = \sum_{i=1}^{N^{rt}}M_i^{cpu}x_{i} + \sum_{j=1}^{N^{be}}M_j^{cpu}y_{j}
\end{equation}
\begin{equation}\label{eq:slo}
\small
t_i + (k_i-x_i)T_{avg} + T \leq \mathcal{T}_{SLO}, \forall i \in [N^{rt}]
\end{equation}
\begin{equation}\label{eq:batch_limit}
\small
\sum_{i=1}^{N^{rt}}x_{i} + \sum_{j=1}^{N^{be}}y_{j} \leq B_{curr}
\end{equation}
\begin{equation}\label{eq:mem_limit}
\small
\sum_{i=1}^{N^{rt}}M^{new}_ix_{i} + \sum_{j=1}^{N^{be}}M^{new}_jy_{j}\leq M^{ept}
\end{equation}

\noindent The BE request serving throughput in (\ref{eq:obj}) is evaluated by the number of generated tokens per unit time. 
In (\ref{eq:T_total}), the cost model $C$ returns the latency of the current iteration, i.e., the time to process the current batch. 
(\ref{eq:l^n}) to (\ref{eq:M^cpu}) give how to derive the inputs $L^n, L^a, M^{cpu}$ for the iteration time lookup. $L^n$ indicates the total input token length 
of all RT and BE requests 
in the batch, $L^a$ is the total length of context (including input and cached tokens) of all the scheduled requests, and $M^{cpu}$ is the total number of checkpointed cache blocks of RT/BE requests scheduled in this batch (store in the CPU memory).  
(\ref{eq:slo}) ensures that the completion time of the current token of each RT request can meet the respective SLO ($\mathcal{T}_{SLO}$ is either the TTFT or TPOT target depending on whether the token is the first or a later one), assuming the remaining $k_i-x_i$ tokens are scheduled in the following iterations consecutively. 
(\ref{eq:batch_limit}) limits the number of selected requests to include in each batch to $B_{curr}$, which is determined by the adaptive batching mechanism. 
(\ref{eq:mem_limit}) guarantees sufficient empty GPU blocks for scheduled RT/BE requests.

\noindent\textbf{Inference cost modeling.}
We build cost models to estimate the LLM serving latency at an iteration. 
The input to LLM inference of a batch includes the hidden state sequences $\in R^{L^n \times D} $ ($D$ is the hidden size) and the context of all requests in the batch.
The model execution latency depends on two shapes: $R^{L^n \times D}$, which is the input shape to none-attention operators like linear projections, activations, and normalization functions~\cite{attention}, as well as collective communication; $R^{L^n \times L^a}$, which is the shape of the attention score matrix, involved in all attention-related operators including batched general matrix multiplication and softmax.
The latency cost model can be described as $\alpha_0 \cdot L^{n} + \alpha_1 \cdot L^{n}L^{a} + \beta$.
Following LLM-PQ~\cite{llm-pq}, we perform offline profiling to generate dummy sequences with different lengths and KV cache sizes, collecting the execution latencies to fit linear models for prefill and decode requests separately.
Sequence lengths are sampled at exponential intervals of 2, with the maximum value being the product of the LLM context length and the maximum batch size.
For batch execution time querying, we separate prefill and decode requests within the current batch, then map $L^n$ of all prefill requests to cost models of prefill-phase model execution ($L^a$ always equals $L^n$ as no tokens are cached yet), and map $L^n$ and $L^a$ to cost models of decode-phase model execution. Finally, we sum up the execution time of requests in these two phases in the batch as $t_{exec}$. 

As scheduled requests may have checkpointed KV cache
in CPU memory (Sec.~\ref{sec:kv}), the potential swapping time from CPU memory to GPU memory also influences the iteration time. We conduct memory bandwidth profiling and develop a linear model to estimate the swapping time $t_{swap}$ on $M^{cpu}$. 
With asynchronous memory operations, the memory swapping time overlaps with the model execution time, and the iteration time can be computed as $\max(t_{exec}, t_{swap})$. 

\subsection{Hybrid Scheduling Algorithm}\label{sec:algorithm}

\begin{figure}[!t]
  \centering
  \includegraphics[width=\linewidth]{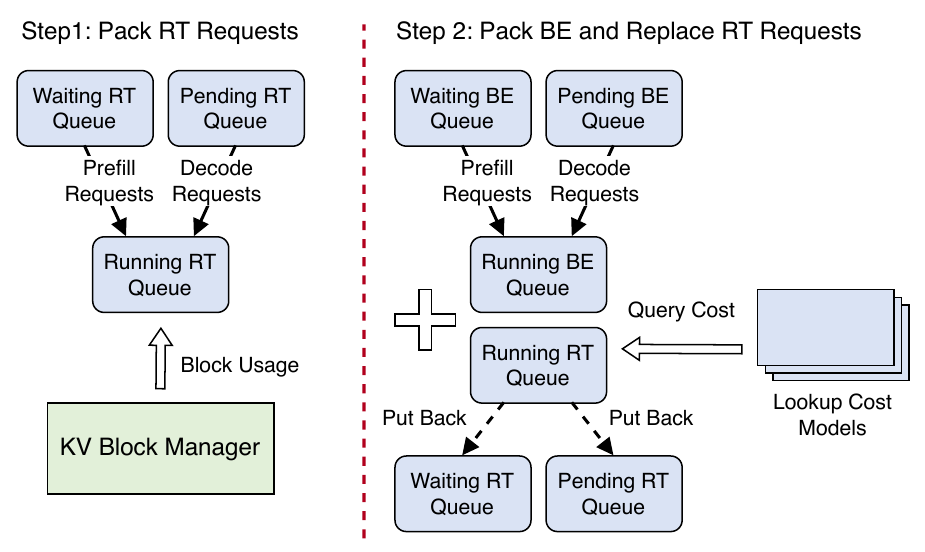}
  \caption{Request scheduling process of the priority-based packing algorithm.
  }
  \label{fig:schedule_process}
\end{figure}

Problem (\ref{eq:obj}) 
can be viewed as a variant of the 1D bin packing problem, where the maximum batch size is the bin capacity. Even ignoring the overhead of estimating the iteration time, there can be up to $\sum_{b_i=1}^{B_{curr}} \binom{N^{rt}+N^{be}}{b_i}$ combinations of requests to schedule, making the problem expensive to solve in runtime. We design a greedy priority-based packing algorithm to efficiently derive scheduling decisions for each iteration.

\noindent\textbf{Main idea.} 
As illustrated in Fig.~\ref{fig:schedule_process}, 
the scheduler first retrieves RT requests in ascending order of their remaining time to meet the token latency SLOs (TTFT or TPOT, depending on each request's prefill or decode phase). 
Next, the scheduler adds BE requests (in ascending order of CPU-to-GPU KV cache swapping overhead) to the running queue if the batch is not full, or replaces lower-priority RT requests (with more remaining time to SLO targets) with BE requests that introduce less swapping overheads.
The scheduler schedules as many requests as possible, that do not exceed the current batch size and ensures that the estimated iteration time of the batch does not exceed the residual time to the SLO deadline of the most urgent RT request (i.e., the one closest to its deadline does not fail).
In this manner, the scheduler can adapt system resources to dynamic workloads by flexibly determining the actual batch size for each iteration, based on the remaining time budget. After deciding the requests to schedule, the RT and BE running queues are merged to form a batch, which is dispatched to the executors. 

\begin{algorithm}[!t]
\caption{Priority-based Packing Algorithm for RT and BE request scheduling at each iteration}\label{alg:pack}
\DontPrintSemicolon
\SetNoFillComment

\SetKwInOut{KwInputs}{Input}
\SetKwInOut{KwOutputs}{Output}

\KwInputs{$Q^{rt}$ - RT request queues, $Q^{be}$ - BE request queues, $Mgr$ - KV block manager, $B_{curr}$ - current batch size, $\mathcal{T}_{SLO}$ - SLO target of RT request} 
\KwOutputs{$RT\_ready$ - scheduled RT requests, $BE\_ready$ - scheduled BE requests}

$RT\_ready \leftarrow$ []\;

\bluecomment{Sort RT requests based on remaining time}\;

\textbf{sort} $Q^{rt}$ \textbf{according to} $\mathcal{T}_{SLO} - t_i - (k_i-1)T_{avg}$ in ascending order\;

$T_{min}^{res} \leftarrow$ $\mathcal{T}_{SLO} \ - $ \textit{GetMinResidualTime}($Q^{rt}$)\;

\While{$\exists \ req^{rt} \in Q^{rt}$}{
\If{ Size($RT\_ready$) == $B_{curr}$}{
\textbf{break}}
\If{\textbf{not} FindPreemptBlock($req^{rt}$, $Mgr$)}{
\textbf{break}}
$cand\_reqs \leftarrow RT\_ready + req^{rt}$\;

\If{$C(cand\_reqs) >T_{min}^{res}$}{
\textbf{break}
}
\textbf{front pop} $req^{rt}$ \textbf{from} $Q^{rt}$\;

$RT\_ready \leftarrow RT\_ready \cup \ req^{rt}$\;
}
$BE\_ready \leftarrow$ []\;

\bluecomment{\textit{Sort BE requests based on $M_j^{cpu}$}}\;

\textbf{sort} $Q^{be}$ \textbf{according to} $M_j^{cpu}$ in ascending order\;

\While{$\exists \ req^{be} \in Q^{be}$}{
\If{\textbf{not} FindBlock($req^{be}$, $Mgr$)}{
%
%
\textbf{break}\;
}

\bluecomment{\textit{First try packing}}\;

\If{$Size(RT\_ready + BE\_ready + req^{be}$) $\le$ $B_{curr}$ 
\textbf{and} $C(RT\_ready + BE\_ready + req^{be}) \le T_{min}^{res}$}{
$BE\_ready \leftarrow BE\_ready \cup \ req^{be}$\;
}
\Else{
\bluecomment{\textit{Then try replacing}}\;

\textbf{back pop} $req^{rt}$ \textbf{from} $RT\_ready$\;

\If{$Size(RT\_ready + BE\_ready + req^{be}$) $\le$ $B_{curr}$ \textbf{and} $C(RT\_ready + BE\_ready + req^{be}) \le T_{min}^{res}$}{
\textbf{front insert} $req^{rt}$ \textbf{to} $Q^{rt}$\;

$BE\_ready \leftarrow BE\_ready \cup \ req^{be}$\;
}
\Else{
$RT\_ready \leftarrow RT\_ready \cup \ req^{rt}$\;

\textbf{break}\;
}
}
\textbf{front pop} $req^{be}$ \textbf{from} $Q^{be}$\;
}
\end{algorithm}

\noindent\textbf{Algorithm.} The request scheduling algorithm is given in Alg.~\ref{alg:pack}.
We first order requests in the RT queues (including pending and waiting queues) altogether by their remaining time to the respective token generation deadlines in ascending order and decide the smallest residual time $T^{res}_{min}$ (lines 3-4). We keep adding the most urgent RT requests from the queues to the batch until the batch size becomes $B$, no empty and preemptible cache blocks (determined by the function \textit{FindPreemptBlock}, which obeys the rule in Sec.~\ref{sec:kv}) exist in the system, or we find that $T^{res}_{min}$ can not be satisfied
(lines 5-15).  
Then the BE queues are sorted by the number of checkpointed slots in ascending order (line 17). Each time, we select the BE request with the least potential memory swapping overhead and try to add the BE request to the batch or replace the least urgent RT request with the BE request (lines 18-33). 

\begin{figure}[!t]
  \centering
  \includegraphics[width=\linewidth]{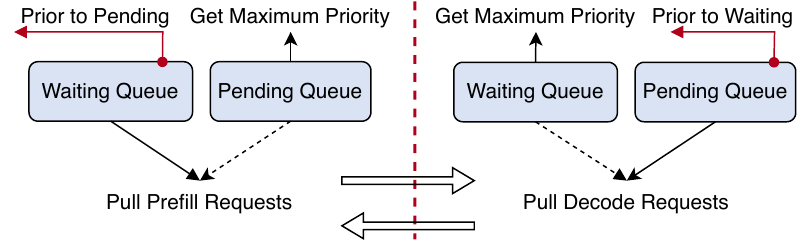}
  \caption{Queues polling to pull RT requests in order.}
  \label{fig:poll}
  \vspace{-5mm}
\end{figure}

\noindent\textbf{Request pulling optimization.}
When pulling RT requests, Alg.~\ref{alg:pack} requires sorting requests across different state queues (pending and waiting) at each iteration.
Naive implementation by merging different queues before sorting and splitting them back after the two-step request selection introduces non-negligible overheads, especially when there exist numerous concurrent requests.
To mitigate scheduling overhead, we opt for partial sorting and continuous queue polling.
The key insight is that the waiting queue is inherently ordered (by request arrival time).
Requests closer to the head of the waiting queue have longer elapsed time.
We only need to sort the pending queue and then do alternative queue polling. As depicted in Fig.~\ref{fig:poll}, we keep retrieving requests of the maximum priority from one queue and pulling requests that are more prioritized from the other.

\noindent\textbf{Adaptive Batch Sizing.}
Before selecting requests from RT and BE queues, the current batch size $B_{curr}$ should be determined to strike a balance between low RT request latency and high BE request throughput.
Drawing inspiration from the window size adjustment mechanism in TCP Congestion Control, we propose a heuristic to dynamically adapt the batch size based on the runtime SLO attainments of RT requests.
Our approach operates as follows:

$\triangleright$ \textit{Initialization:} 
We start with a conservative batch size, e.g., $B_{curr} \leftarrow B_{base} = 128$.

$\triangleright$ \textit{gradual Expansion:} 
At each iteration, if the scheduler does not trigger the early-return in lines 11-12 of Alg.~\ref{alg:pack}, which means that no RT requests are detected, we increase the batch size: $B_{curr} \leftarrow B_{curr} \cdot 2$ to include more BE requests for batching.

$\triangleright$ \textit{Aggressive Backoff:} 
Once the estimated iteration time surpasses the residual timev $T_{min}^{res}$, we apply an aggressive penalty by resetting the current batch size $B_{curr}$ back to $B_{base}$.

\section{Bidirectional KV Cache Management}\label{sec:kv}
Since request scheduling is constrained by available memory in the system for KV caching, it is crucial to efficiently allocate and manage memory resources (for KV cache storage) among concurrent requests.
\sys{}'s KV cache management adopts a bidirectional storage layout, along with the block preemption and lazy checkpointing mechanisms to flexibly share and allocate GPU memory for RT/BE requests.

\begin{figure}[!t]
  \centering
  \includegraphics[width=\linewidth]{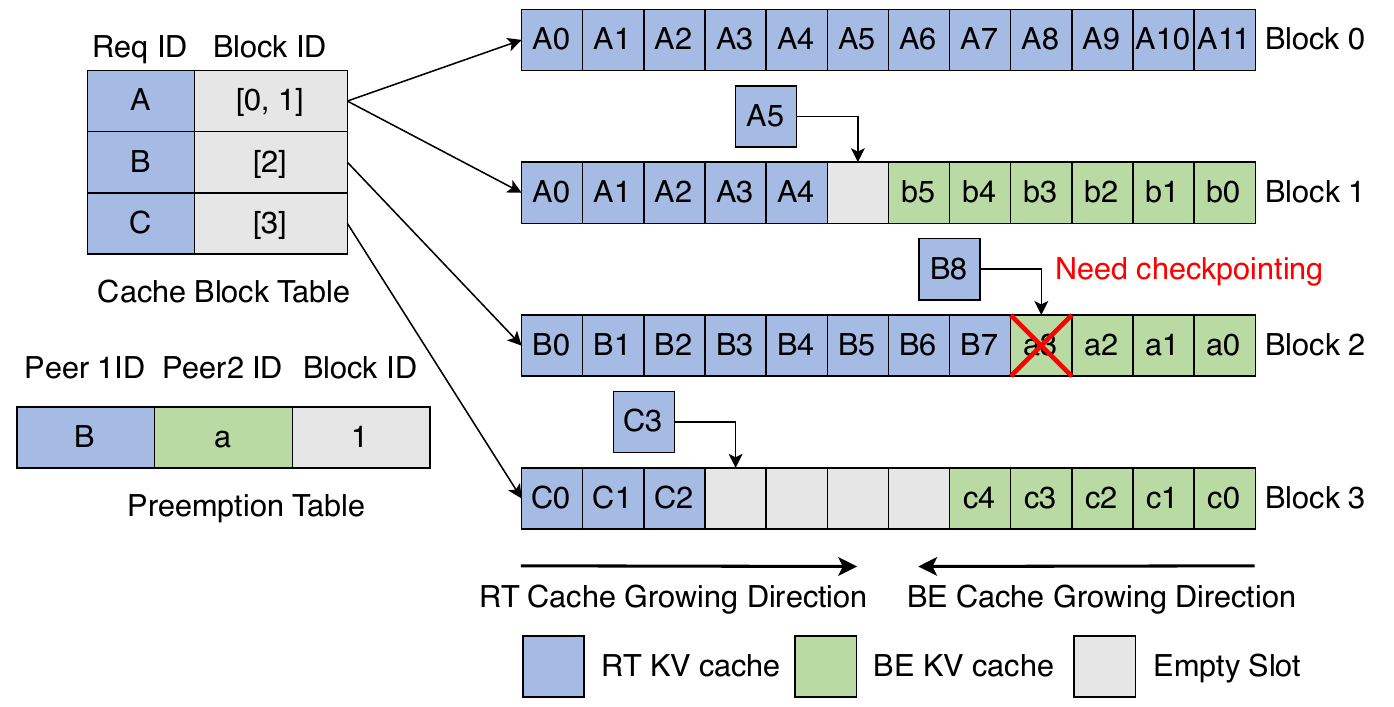}
  \caption{KV cache storage layout for RT and BE requests.
  We only show the RT cache block table for clarity of the figure.
  }
  \label{fig:layout}
  \vspace{-5mm}
\end{figure}

\subsection{Block Structure}\label{sec:block_structure}

KV cache storage renders a major memory bottleneck in LLM serving \cite{vllm}. To manage the KV cache of RT/BE requests, we organize GPU memory for KV cache storage into blocks and each block includes several slots, for storing tokens' KV tensors. Each block can be used for KV cache storage of one RT request and one BE request, with a bidirectional storage layout for sharing the block between two requests: KV cache of the RT request occupies memory slots from the left to the right in the block while that of the BE request in the opposite direction. As depicted in Fig.~\ref{fig:layout}, RT requests A, B, and C occupy slots starting from the left end in blocks 0, 1, 2, and 3, respectively, while BE requests b, a, and c use slots from the right end of the respective blocks.
An RT or BE request can occupy more than one block if its KV cache is large.

\subsection{Block Management}\label{sec:block_manage}

\noindent\textbf{Allocation and preemption.}
When a BE request is scheduled in an iteration, function \textit{FindBlock} in Alg.~\ref{alg:pack} implements the allocation logic: (i) if it is a prefill request whose KV cache has not been stored, we find available GPU blocks for it according to the maximum number of empty slots in each block; (ii) if the request's KV cache has been stored in some GPU blocks, we continue growing its KV cache in the last block (if the empty slots of this block are not used up by itself)
or start occupying a new block (otherwise).
For RT requests, if no GPU blocks are available, \textit{FindPreemptBlock} helps them preempt BE cache blocks and store the KV tensors in the opposite direction.
A straightforward heuristic is adopted to always preempt BE cache blocks with the most empty slots. 
This approach ensures that urgent RT requests will not fail to be scheduled due to insufficient GPU blocks with a large number of concurrent requests.
In extreme cases where no BE-only cache blocks are available for preemption by newly scheduled urgent RT requests, we progressively drop requests: first BE, then RT, based on the remaining time, and swap out their cache blocks.
Once a request generates all tokens, its occupied KV cache slots are released. 

\noindent\textbf{Lazy checkpointing.} 
Within a block, if sufficient empty slots are still available, both RT and BE requests can consider the block as exclusively theirs and safely store their most recent KV tensors.
When all empty slots are exhausted 
during inference, 
only the KV tensors of a specific request that are about to be overwritten by its peer 
need to be checkpointed in CPU memory, instead of the request's entire KV cache~\cite{vllm, fastserve}, and this block is set to preemption status. Those missing KV tensors are swapped back into GPU memory when the request is scheduled in future iterations.
For example, in Fig.~\ref{fig:layout}, KV tensors a3 of BE request \emph{a} is preempted by B8 of RT request \emph{B}, and a3 needs to be checkpointed to the CPU memory.
This design removes unnecessary swapping operations, reducing memory overheads.  
To indicate slot preemption of the blocks, the KV block manager maintains a preemption table, which records the indices of all blocks where slot preemption happened and the indices of RT/BE requests sharing those blocks.
When serving requests with preempted KV cache slots, the scheduler checks the preemption table and generates memory-swapping instructions for the executors to perform corresponding operations.
Before dispatching scheduled requests to the executors, all swapping operations are consolidated and conducted asynchronously to enhance overall efficiency.

\begin{table}[!t]
\caption{Workload statistics. Unit: token.}\label{tab:workload}
\centering
\resizebox{\linewidth}{!}{
    \begin{tabular}{ccccc}
        \toprule
         Type  & Avg. Prompt & Std. Prompt & Avg. Output & Std. Output \\
         \hline
         RT-SharedGPT & 222.76 & 256.36 & 234.51 & 268.50\\
         RT-LMSYS-CHAT-1M & 90.50 & 148.52 & 237.04 & 228.21\\
         BE-Synthetic & 784.14 & 151.17 & 80.96 & 28.23 \\
         \bottomrule
    \end{tabular}}
    \vspace{-3mm}
\end{table}

\section{Implementation}

We implement \sys{} using 8.5k LoC in Python and 2k LoC in C++/CUDA. 
The RPC messages between the engine and the executors are handled with Ray~\cite{ray}, and the tensor transmission is implemented via NCCL~\cite{nccl}. We implement LLMs including OPT~\cite{opt} and Llama~\cite{llama} with the operators provided by PyTorch~\cite{pytorch} and Xformers~\cite{xformers}, aligning our implementation with 
Hugging Face~\cite{huggingface} to load pre-trained model weights.

To serve the scheduled batch, which includes different requests from different phases in an iteration, we concatenate all prefill requests in the batch along the sequence length dimension, followed by the decode requests.
The shapes of individual requests are maintained in a length table.
An additional attention-kernel dispatch function is used to identify the boundary between prefill and decode requests and dispatch the batched prefill and decode requests to their respective attention kernels.
This optimization reduces the CUDA launching overhead associated with Selective Batching~\cite{orca} in Orca and allows scheduling regardless of the stages of requests, which are not available in vLLM.

To enable the bidirectional block layout of the KV cache, we modify the PagedAttention kernel~\cite{vllm}.
Our modifications include a binary direction table, which possesses an identical shape to the block tables. The purpose of this table is to guide the physical block offset for each warp, indicating whether threads should iterate from left or vice versa. 
Furthermore, we leverage PTX (Parallel Thread Execution) instructions to invert the order of elements within the loaded value vector, whenever the flag of direction is evaluated to be true. For checkpointing, we use different CUDA streams to handle asynchronous data movement events across GPU and CPU memory during runtime and customize CUDA kernels to efficiently fetch and store KV caches in non-continuous GPU memory.

\begin{figure*}[!t]
    \centering
     \begin{subfigure}[b]{0.33\linewidth}
     \centering
     \includegraphics[width=\linewidth]{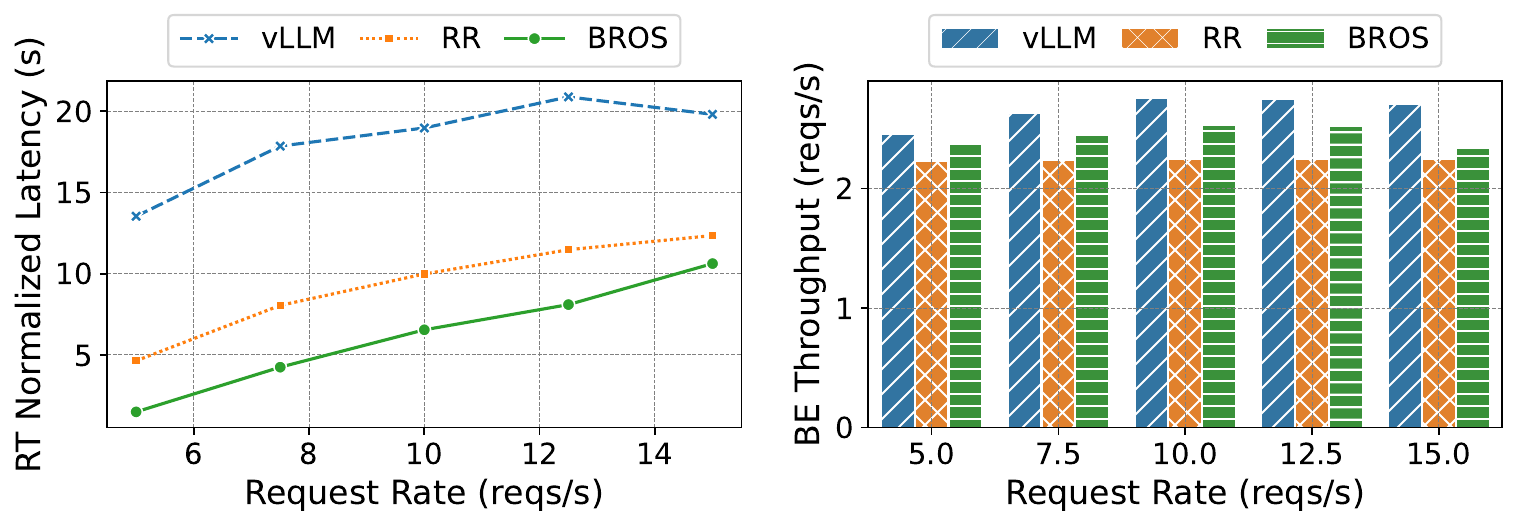}
     \caption{OPT-13B, ShareGPT.}
     \label{fig:opt-13b-sharedgpt_latency_thp}
    \end{subfigure}
      \hfill
     \begin{subfigure}[b]{0.33\linewidth}
         \centering
         \includegraphics[width=\linewidth]{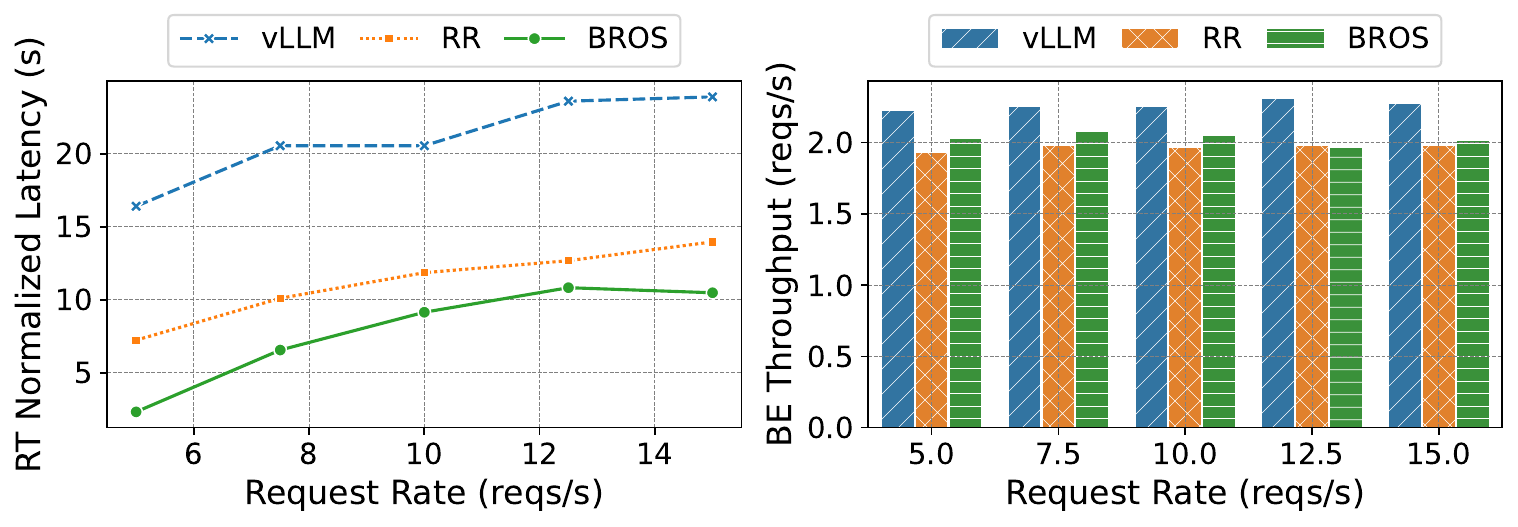}
         \caption{OPT-30B, ShareGPT.}
         \label{fig:opt-30b-sharedgpt_latency_thp}
     \end{subfigure}
     \hfill
     \begin{subfigure}[b]{0.3\linewidth}
         \centering
         \includegraphics[width=\linewidth]{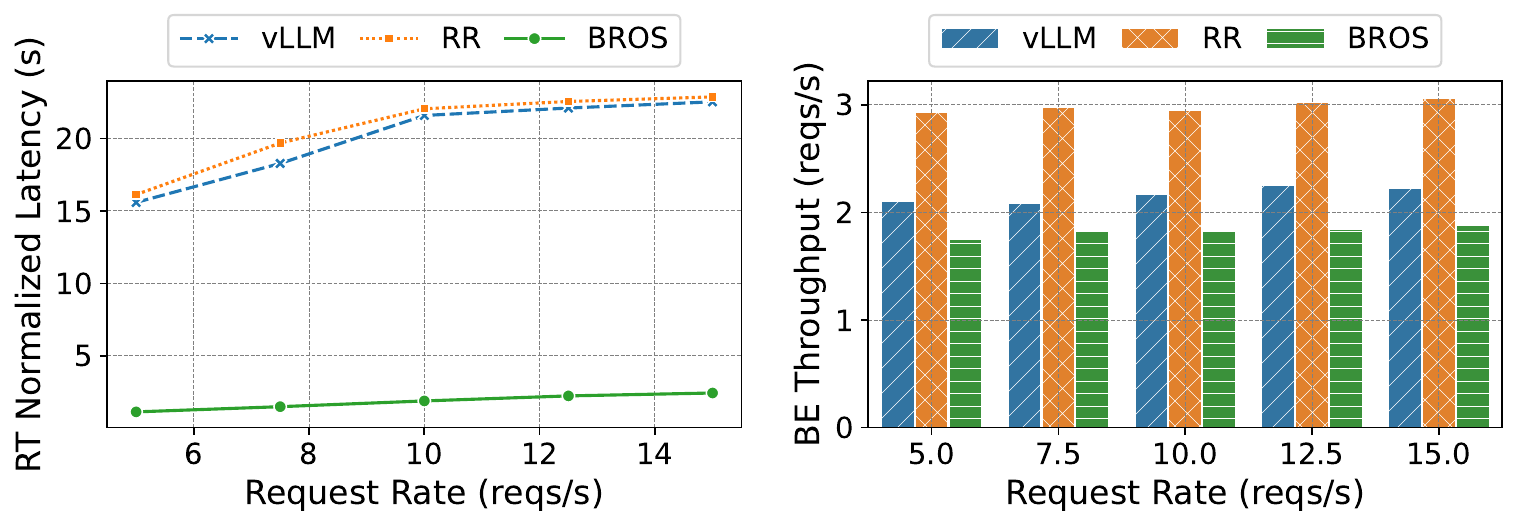}
         \caption{Llama-3-70B, ShareGPT.}
         \label{fig:llama-70b-sharedgpt_latency_thp}
     \end{subfigure}
     \hfill
     \begin{subfigure}[b]{0.33\linewidth}
     \centering
     \includegraphics[width=\linewidth]{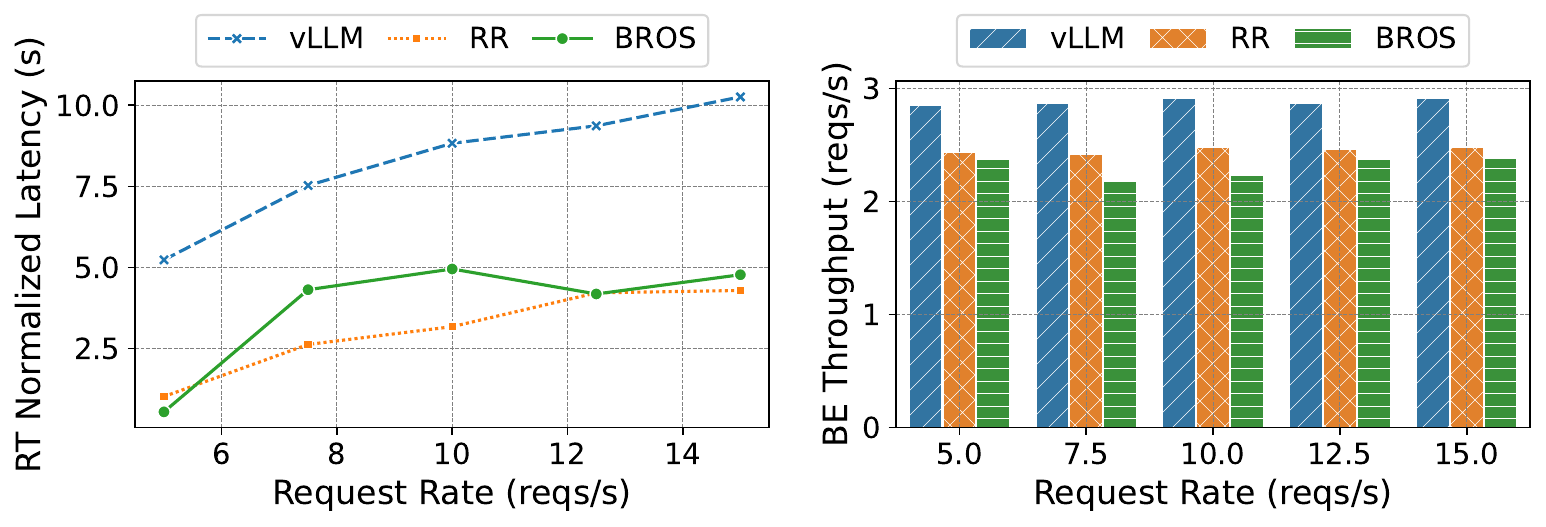}
     \caption{OPT-13B, LMSYS-CHAT.}
     \label{fig:opt-13b-lmsys_chat_latency_thp}
    \end{subfigure}
      \hfill
     \begin{subfigure}[b]{0.33\linewidth}
         \centering
         \includegraphics[width=\linewidth]{images/experiments/RT_BE_main_1/13B_lmsys_chat.pdf}
         \caption{OPT-30B, LMSYS-CHAT.}
         \label{fig:opt-30b-lmsys_chat_latency_thp}
     \end{subfigure}
     \hfill
     \begin{subfigure}[b]{0.3\linewidth}
         \centering
         \includegraphics[width=\linewidth]{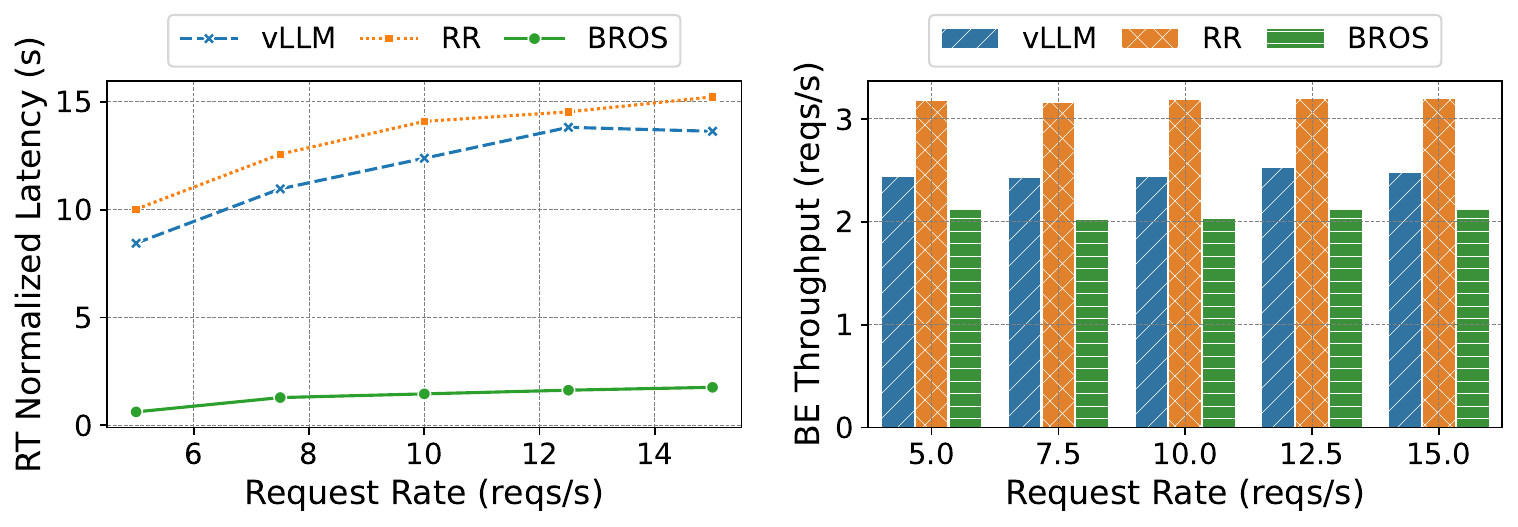}
         \caption{Llama-3-70B, LMSYS-CHAT.}
         \label{fig:llama-70b-lmsys_chat_latency_thp}
     \end{subfigure}
    \caption{Comparison of latency and throughput among \sys{}, vLLM, and RR on RT/BE hybrid workloads.}
    \label{fig:latency_throughput}
\end{figure*}

\begin{figure*}[!t]
    \centering
     \begin{subfigure}[b]{0.33\linewidth}
     \centering
     \includegraphics[width=\linewidth]{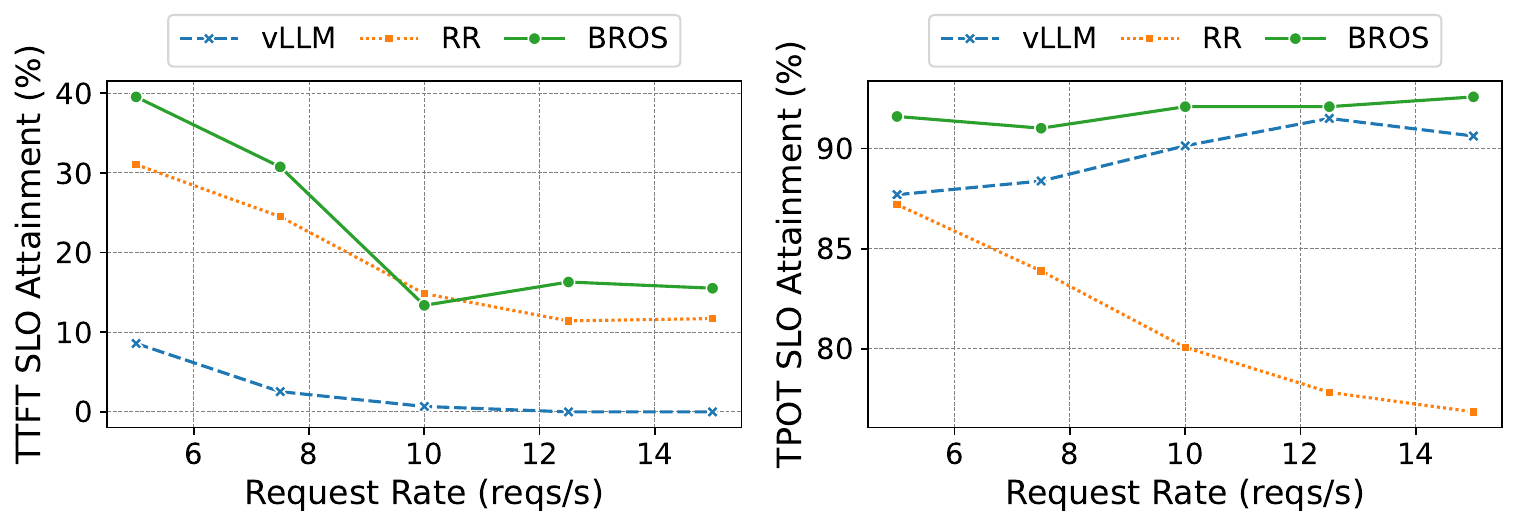}
     \caption{OPT-13B, ShareGPT.}
     \label{fig:opt-13b-sharedgpt_slo}
    \end{subfigure}
     \hfill
     \begin{subfigure}[b]{0.33\linewidth}
         \centering
         \includegraphics[width=\linewidth]{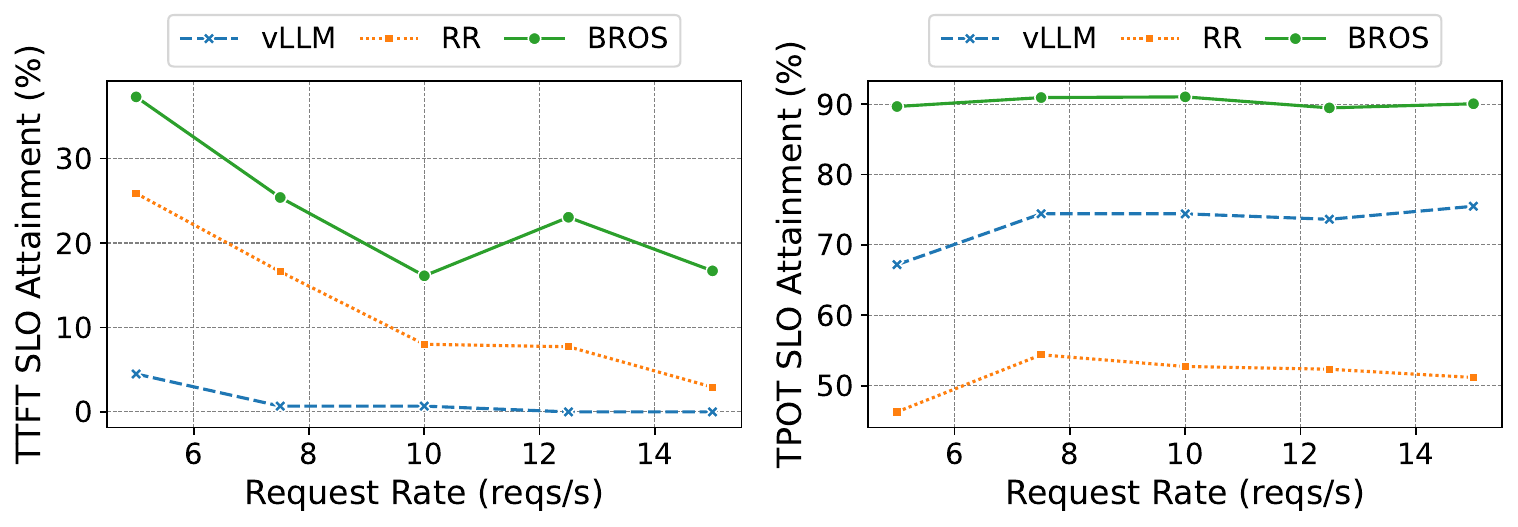}
         \caption{OPT-30B, ShareGPT.}
         \label{fig:opt-30b-sharedgpt_slo}
     \end{subfigure}
     \hfill
     \begin{subfigure}[b]{0.33\linewidth}
         \centering
         \includegraphics[width=\linewidth]{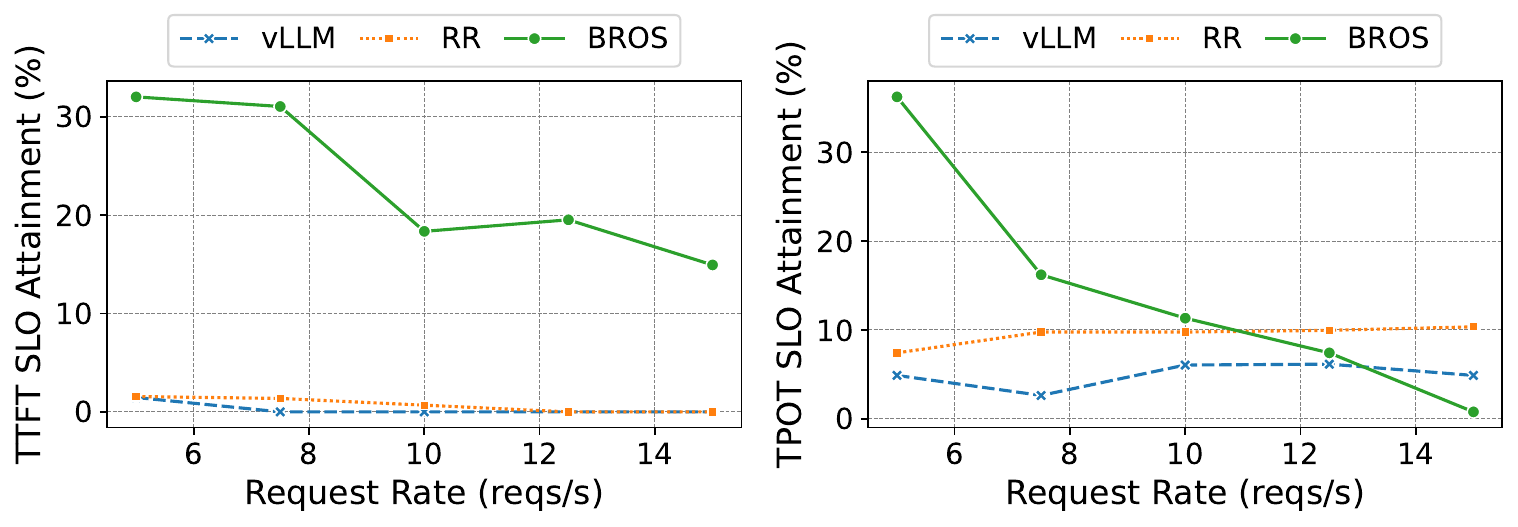}
         \caption{Llama-3-70B, ShareGPT.}
         \label{fig:llama-70b-sharedgpt_slo}
     \end{subfigure}
     \hfill
     \begin{subfigure}[b]{0.33\linewidth}
     \centering
     \includegraphics[width=\linewidth]{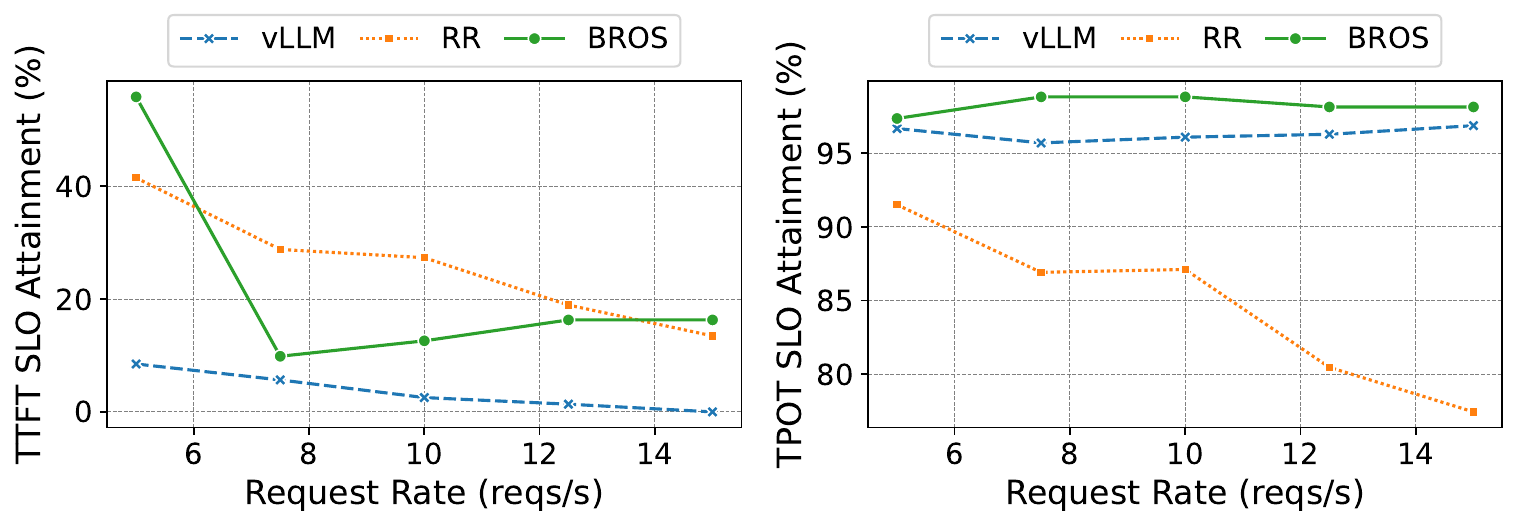}
     \caption{OPT-13B, LMSYS-CHAT.}
     \label{fig:opt-13b-lmsys_chat_slo}
    \end{subfigure}
     \hfill
     \begin{subfigure}[b]{0.33\linewidth}
         \centering
         \includegraphics[width=\linewidth]{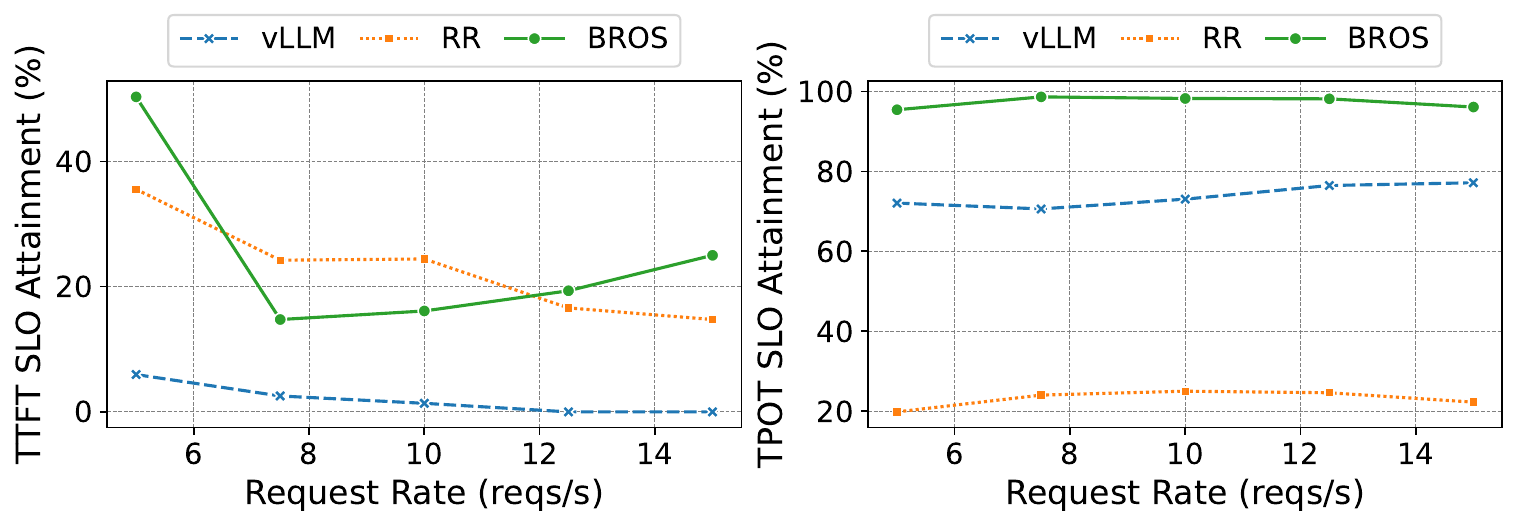}
         \caption{OPT-30B, LMSYS-CHAT.}
         \label{fig:opt-30b-lmsys_chat_slo}
     \end{subfigure}
     \hfill
     \begin{subfigure}[b]{0.33\linewidth}
         \centering
         \includegraphics[width=\linewidth]{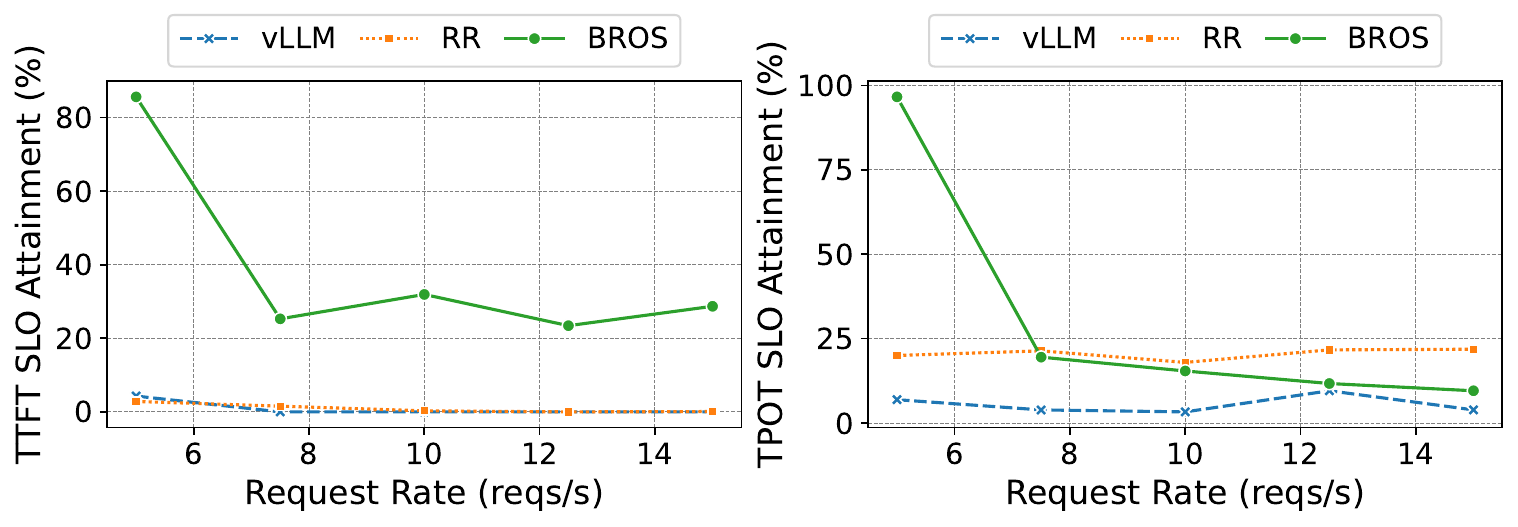}
         \caption{Llama-3-70B, LMSYS-CHAT.}
         \label{fig:llama-70b-lmsys_chat_slo}
     \end{subfigure}
    \caption{Comparison of SLO attainments among \sys{}, vLLM, and RR on RT/BE hybrid workloads.}
    \label{fig:slo_attainment}
\end{figure*}

\section{Evaluation}\label{sec:exp}
\subsection{Experimental Setup}

\noindent\textbf{Models and testbed.} We run OPT~\cite{opt} models with 13B, 30B parameters, and Llama-3~\cite{llama3} with 70B parameters. For experiments on OPT models, we use a server (Ubuntu 20.04LTS) with 4 NVIDIA A100-SXM4-40GB GPUs interconnected by NVLink. For experiments on the Llama-3 model, we use a server (Ubuntu 20.04LTS) with 8 NVIDIA A100-SXM8-80GB GPUs connected by NVSwitch. All models are served with tensor parallelism with parallel degrees of OPT-13B, OPT-30B, and Llama-70B being 2, 4, and 8, respectively. To ensure a fair comparison, we employ greedy sampling as the decoding algorithm
for all LLMs, compelling the models to generate a specified number of tokens.


\noindent\textbf{Traces.} 
Following vLLM~\cite{vllm}, we synthesize RT requests based on the ShareGPT~\cite{sharegpt} dataset and LMSYS-CHAT-1M datasets~\cite{lmsys}, both of which contain real-world user conversations with LLM services.
Since no representative BE request dataset is available, we follow FlexGen~\cite{flexgen} to use synthetic workloads. 
All prompt and output lengths for BE requests are sampled from uniform distributions with ranges (512, 1024) and (32, 128).
Similar to previous works~\cite{orca, vllm, distserve, llumnix}, we generate the arrival times of RT requests using Poisson distribution with different request rates, while BE requests are periodically submitted in a batch manner when previous batched requests are completed~\cite{batch-api}.
Further details of the workloads are in Table~\ref{tab:workload}.

\noindent\textbf{Baselines.}
We compare \sys{} with three baselines: (1) \textit{vLLM}, the SOTA LLM serving system optimized for high-throughput text generation; (2) \textit{TGI}~\cite{TGI}, an production-grade LLM serving system used in LLM services at Hugging Face, to support applications like Hugging Chat; (3) \textit{\sys{}-Round-Robin} (referred to as RR), where we replace the scheduling algorithm of \sys{} with a Round-Robin schedule described in Sec.~\ref{sec:llm_chan} and remove the bidirectional KV cache management.
We do not include a comparison with offline LLM serving systems, such as FlexGen~\cite{flexgen} and LLM-PQ~\cite{llm-pq}.
This is because they do not implement continous batching~\cite{orca}, thereby introduce substantial overhead when co-serving RT requests.
For fair comparisons, we set the KV cache block size to 16 for vLLM, RR, and \sys{}, and the total KV cache capacity per GPU is obtained by profiling the maximum available GPU memory for the KV cache.

\noindent\textbf{Metrics.} 
We measure the following metrics on RT requests: (1) \textit{normalized latency}~\cite{orca, vllm}, computed as the mean of each RT request's end-to-end latency divided by the number of its output tokens; (2) \textit{time to first token} (TTFT), representing the latency for each RT request to return the first token after submitting its prompt; (3) \textit{time per output token} (TPOT), indicating the average latency for generating subsequent tokens.
We set the SLO for TPOT to be 0.2s, 
which is a widely accepted latency for one forward pass of a DNN~\cite{dvabatch, nexus}.
For TTFT, considering that prefill processing takes longer~\cite{splitfuse, distserve}, we set its SLO to be 0.4s.
We use throughput (measured in requests per second) to evaluate the serving performance of BE requests.
We show the average normalized latency and SLO attainments, computed as the mean of each request's normalized latency and SLO attainment.
For all experiments, we evaluate the systems with 10-minute traces.

\begin{figure*}[!t]
    \centering
     \begin{subfigure}[b]{0.49\linewidth}
     \centering
     \includegraphics[width=\linewidth]{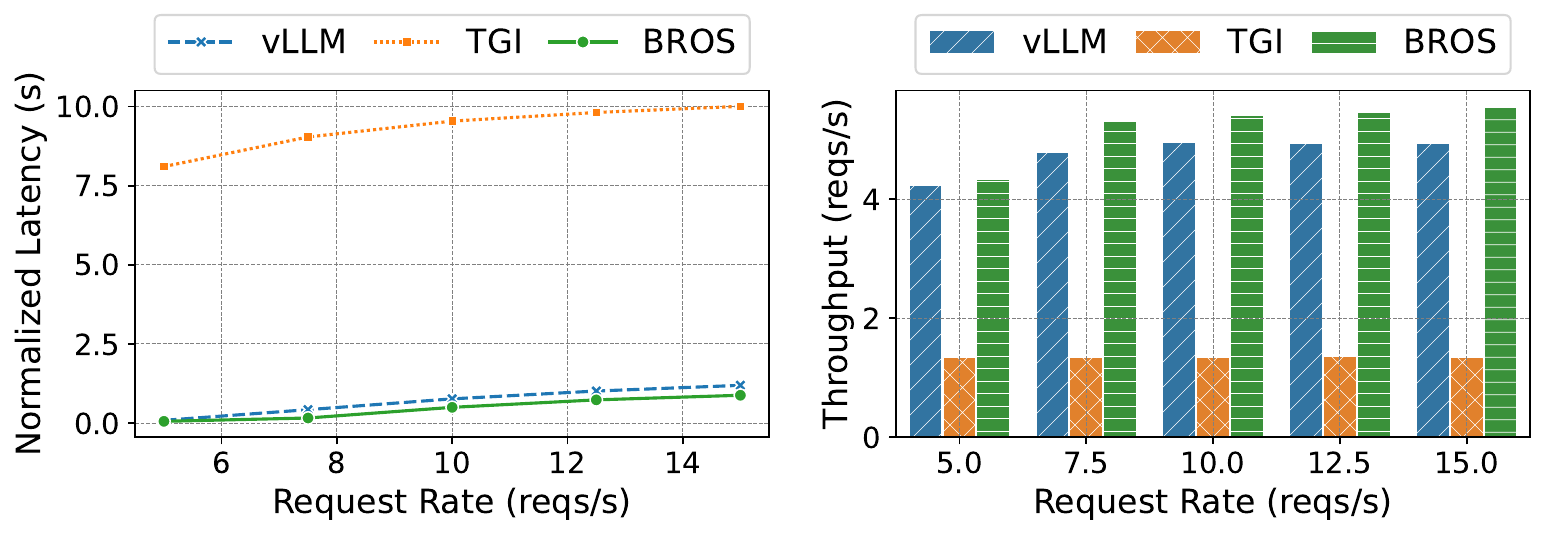}
     \caption{OPT-13B, 2GPUs, SharedGPT.}
     \label{fig:opt-13b-sharedgpt_RT}
    \end{subfigure}
      \hfill
     \begin{subfigure}[b]{0.49\linewidth}
         \centering
         \includegraphics[width=\linewidth]{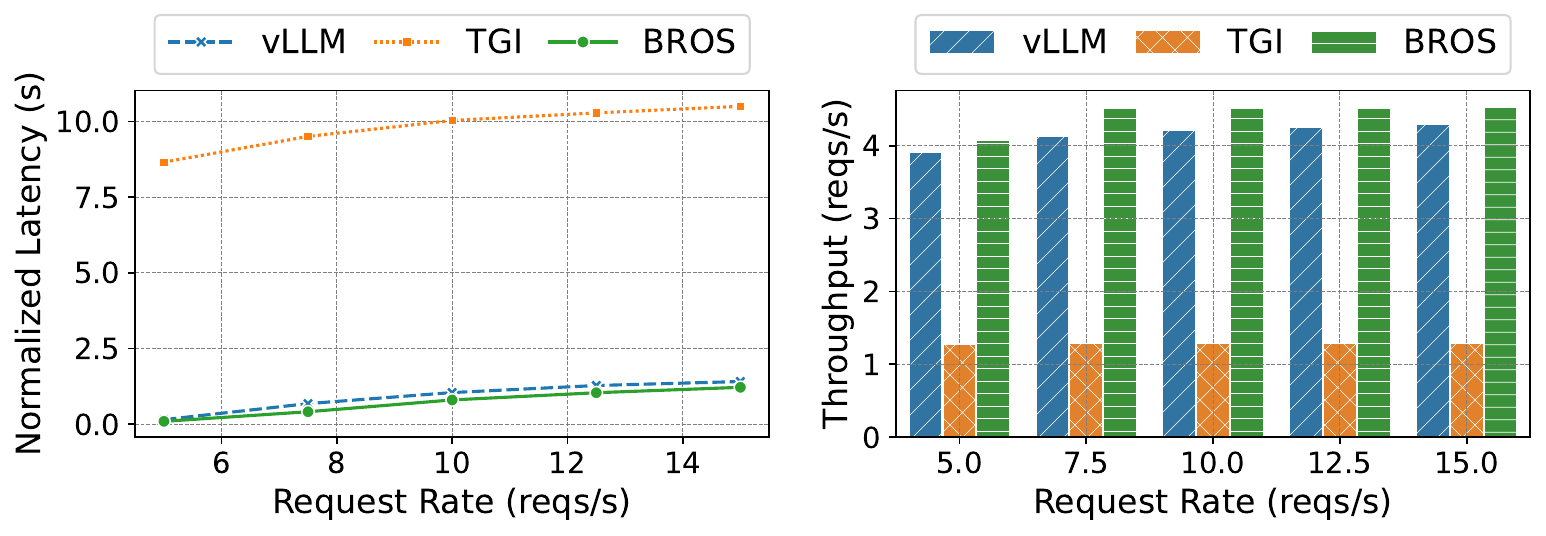}
         \caption{OPT-30B, 4GPUs, SharedGPT.}
         \label{fig:opt-30b-sharedgpt_RT}
     \end{subfigure}
    \caption{Comparison of latency and throughput among \sys{}, vLLM, and TGI on pure RT workloads.}
    \label{fig:rt_only}
\end{figure*}

\subsection{Striking Better Trade-off on Hybrid RT/BE Workloads}\label{sec:exp_main}

We evaluate the end-to-end performance of \sys{} with vLLM and RR on hybrid RT and BE requests.
Performance comparisons on different metrics are demonstrated in Fig.~\ref{fig:latency_throughput} and Fig.~\ref{fig:slo_attainment}.

\subsubsection{Latency and Throughput}
Figure~\ref{fig:opt-13b-sharedgpt_latency_thp} to Fig.~\ref{fig:llama-70b-sharedgpt_latency_thp} show that when RT requests are sampled from the SharedGPT dataset,
\sys{} demonstrates reductions of 74.20\% and 54.31\% in average normalized latency, as compared to vLLM and RR at various request rates.
Simultaneously, \sys{} maintains high throughput for BE requests, with only an 11.29\% reduction compared to vLLM and an 8.80\% reduction compared to RR.
The same benefits are also observed on the LMSYS-CHAT-1M datasets (Fig.~\ref{fig:opt-13b-lmsys_chat_latency_thp} to Fig.~\ref{fig:llama-70b-lmsys_chat_latency_thp}).
In this case, \sys{} achieves an average normalized latency reduction of 69.61\% and 25.77\% for RT requests, with a 17.67\% and 15.18\% reduction in throughput compared to vLLM and RR, respectively.
In particular, Fig.~\ref{fig:llama-70b-sharedgpt_latency_thp} and Fig.~\ref{fig:llama-70b-lmsys_chat_latency_thp} show that, though sacrificing BE throughput, \sys{} consistently maintains low latency for RT requests even under an extremely high request rate (15 req/s) and with a huge model size (70B).
The performance gains can be attributed to our scheduling mechanism, which identifies different urgent RT requests at each iteration and assigns them higher priorities, surpassing the reliance on arrival orders or fair scheduling with BE requests.
The second step of the scheduling algorithm focuses on replacing non-urgent RT requests and maximizing the scheduling of BE requests.
Furthermore, the bidirectional KV cache management of \sys{} allows RT to share KV cache blocks of BE requests, ensuring sufficient memory for urgent RT requests. 
As a result, \sys{} strikes a better trade-off by sacrificing a small proportion of BE requests' throughput in exchange for substantial improvements in RT requests' latency.



\subsubsection{SLO Attainments}
On the SharedGPT dataset, \sys{} demonstrates 36.38$\times$ and 1.73$\times$ improvements in TTFT and TPOT SLO attainments over vLLM.
Compared to RR, \sys{} achieves improvements of 11.75$\times$ and 1.47$\times$ in TTFT and TPOT. 
On LMSYS-CHAT-1M, \sys{} achieves improvements of 21.42$\times$ and 2.62$\times$ over vLLM, and 14.23$\times$ and 2.28$\times$ over RR, respectively.
These improvements are due to the designs of \sys{}'s scheduling targets for RT requests, which consider the quick response and high context rate simultaneously when making decisions.
Additionally, the adaptive batch sizing mechanism assists in regulating the maximum number of batched requests based on feedback from RT request SLO attainment.
No baseline method effectively balances TTFT and TPOT SLO attainments.
vLLM achieves an average TTFT SLO attainment of only 1.28\% and fails to meet the TTFT SLOs in 8 out of 15 cases.
RR alleviates this by serving RT and BE requests alternately at each iteration, but the results (10.55\%) are still unsatisfactory.
Regarding TPOT SLO attainment, vLLM is expected to achieve good results since when its running queue is full, all scheduled requests can complete generation without further preemption.
However, without proper memory contention management, later-scheduled RT requests are compelled to release memory due to insufficient available resources; these requests must wait until earlier RT/BE requests finish generation regardless of their urgency.
\sys{} effectively addresses this issue through bidirectional block sharing and block preemption, resulting in higher TPOT attainments.
RR achieves lower average TPOT attainment (47.34\% on SharedGPT and 42.81\% on LMSYS-CHAT-1M) compared to \sys{} (65.49\% on SharedGPT and 75.38\% on LMSYS-CHAT-1M).
This is because RR consistently interrupts the processing of RT requests whenever there are BE requests.
Unlike naive preemptive or static fair scheduling, \sys{} always seeks to balance these two SLO attainments.

\vspace{-3mm}
\subsection{System Microbenchmark}
\label{sec:exp_micro}
We compare \sys{} with vLLM and TGI for the system microbenchmark.
We use the RT requests from Table~\ref{tab:workload}, and eliminate the second part of the priority-based packing algorithm and the bidirectional KV cache management as well.
We evaluate the systems' serving performance with normalized latency and throughput, consistent with those in previous studies~\cite{orca, vllm}. 
Fig.~\ref{fig:rt_only} shows that \sys{} still achieves a 29.36\% reduction in latency and 1.08$\times$ throughput improvement over vLLM. 
Despite scheduling overheads, \sys{} can outperform SOTA systems in end-to-end performance.

\subsection{Ablation Studies}\label{sec:exp_ab}
We conduct ablation experiments to investigate various factors that influence the performance of \sys{}.

\begin{figure}[!t]
  \centering
  \includegraphics[width=0.98\linewidth]{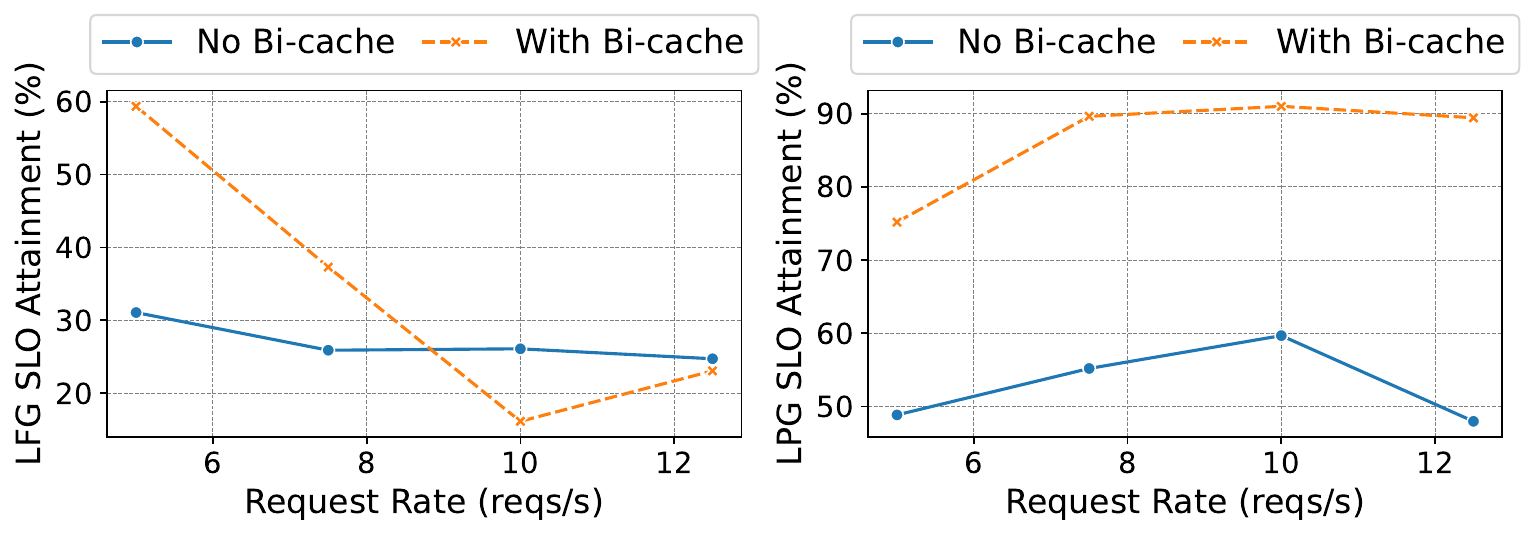}
  \caption{TTFT and TPOT SLO attainment: with and without bi-directional cache management.}
  \label{fig:benefit_bi}
\end{figure}

\subsubsection{Benefits of Cache Management} 
 In Fig.~\ref{fig:benefit_bi}, we compare the serving performance of RT requests with/without the bidirectional KV cache management (denoted as With Bi-cache and No Bi-cache, respectively).
 We serve OPT-30B on hybrid workloads, where 
 the request rate is fixed to 10.
 In the case without bi-directional cache management, we simply perform recomputation for requests failing to acquire cache blocks.
With our designs, TTFT and TPOT SLO attainments are 1.23$\times$ and 1.63$\times$ better than without, which verifies the benefits of our block layout for efficient resource sharing and better SLO attainments of RT requests.

\begin{figure}[!t]
    \centering
     \begin{subfigure}[b]{0.49\linewidth}
     \centering
     \includegraphics[width=\linewidth]{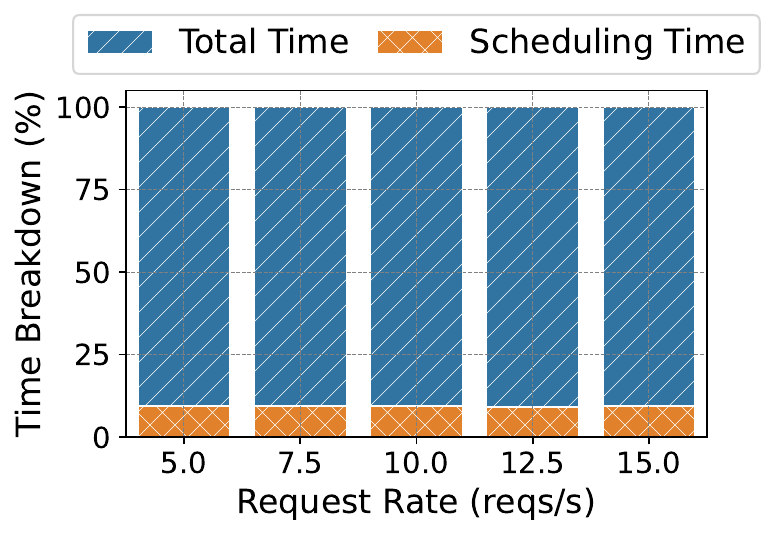}
     \caption{OPT-13B.}
     \label{fig:schedule_overhead_13B}
    \end{subfigure}
     \hfill
     \begin{subfigure}[b]{0.49\linewidth}
         \centering
         \includegraphics[width=\linewidth]{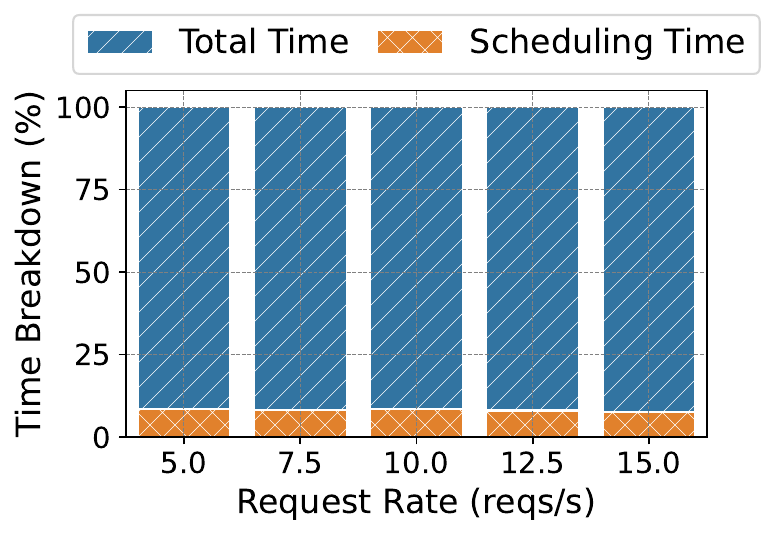}
         \caption{OPT-30B.}
         \label{fig:head_of_line_30B}
     \end{subfigure}
     \caption{Scheduling Overhead.}
      \label{fig:schedule_overhead}
      \vspace{-5mm}
\end{figure}

\subsubsection{Scheduling Overheads} 
We evaluate the overheads of the request scheduling, which involves sorting RT and BE requests and multiple cost model queries at each iteration.
We run OPT-13B and OPT-30B models with hybrid RT and BE requests as in Sec.~\ref{sec:exp_main}.
We measure the time to finish scheduling as well as the total execution time at each iteration, showing the time breakdown.
Fig.~\ref{fig:schedule_overhead} illustrates that the scheduling overhead ranges from 9.12\% to 9.52\% per iteration for OPT-13B and from 7.56\% to 8.42\% for OPT-30B. Such overheads are justified with the gains achieved by \sys{} in terms of reducing RT request latency and improving normalized latency SLO attainment.

\section{Related Works}\label{sec:related_work}

\noindent\textbf{Inference Techniques.}
Some LLM-workload-agnostic techniques are adopted by LLM serving systems to further enhance performance.
FlashAttention~\cite{flashattention} and FlashInfer~\cite{flashinfer} leverage online softmax to optimize implementations of LLM's attention kernels, which improves the computation efficiency.
Flux~\cite{flux} and NanoFlow~\cite{zhu2024nanoflow} adopt
communication-computation overlapping to further hide the communication overheads during LLM inference.
In addition, various distributed inference strategies, such as Tensor Parallelism~\cite{megatron}, Sequence Parallelism~\cite{ring, ulysses}, and Expert Parallelism~\cite{deepep}, are employed by serving systems to scale Dense/Sparse LLM inference.

\noindent\textbf{LLM serving on RT/BE requests.}
Existing works improve LLM serving performance for RT or BE requests but not both together.
For RT requests, Orca~\cite{orca} proposes iteration-level scheduling, which returns finished requests immediately and adds new requests to the current batch. FastServe~\cite{fastserve} supports preemptive scheduling with a skip-join MLFQ, further reducing the average latency. vLLM~\cite{vllm} advocates PagedAttention to reduce memory fragmentation caused by KV cache storage and enable larger batch sizes to improve throughput. Sarathi-Serve~\cite{taming} splits long prompts into chunks and directly adds them into ongoing decode batches without 
a stall, therefore reducing processing latency for requests in the decode phase (Sec.~\ref{sec:llm_pattern}). DistServe~\cite{distserve} decouples prefill and decode phases of LLM computation and assigns them to different GPUs to eliminate interference, reducing the latency in each phase. Llumnix~\cite{llumnix} implements an efficient live migration scheme and reschedules them across multiple model instances to fulfill their different priorities. 
For BE requests, FlexGen~\cite{flexgen} advocates fine-grained tensor swapping and placement algorithms over the memory hierarchy to reduce GPU memory consumption during inference.
LLM-PQ~\cite{llm-pq} designs the layer partition and quantization schemes for running LLM services on heterogeneous clusters.

\noindent\textbf{RT/BE requests co-serving.}
In more general domains, RT/BE requests co-serving systems often explore efficient hardware resource sharing.
Shenago~\cite{shenango} runs a scheduling component, i.e., IOkernel, on a dedicated core to conduct high-frequency (every 5$\mu$s) core reallocation for RT/BE applications in the datacenters, based on an efficient allocation algorithm.
The algorithm detects congestion for each application and selects extra cores for it based on hyper-threading efficiency, cache locality, etc.
ELFEN~\cite{elfen} introduces \textit{principled borrowing} to borrow idle cycles from underutilized SMT cores of RT requests for BE workloads, without interfering with RT requests. 
Ubik~\cite{ubik} takes cache space away from RT requests during their idling for batch workloads and boosts their space when they are active again.  
REEF~\cite{reef} is the SOTA serving system that schedules hybrid RT/BE requests for traditional DNNs~\cite{resnet, vgg}, whose inference passes are fixed.
It devises a rule-based kernel packing heuristic for hybrid serving based on profiled inference latency.
However, these core allocation or kernel packing techniques cannot be transferred to LLM serving, where RT/BE requests, in the context of LLM inference, share the same LLM weight and computation kernels.
Serving hybrid RT/BE requests within the same LLM has not been thoroughly explored by these prior works. 
\section{Conclusion}
This paper presents \sys{}, an LLM serving system for hybrid RT/BE requests. We define novel \emph{token group} based latency metrics to balance quick response and high context rate requirements for RT serving, propose a priority-based packing algorithm for preemptive scheduling of RT/BE requests at the iteration level, achieving a favorable serving trade-off between low latency for RT requests and high throughput for BE requests.
To handle memory contention, we advocate bidirectional KV cache management, which properly shares cache blocks between concurrent RT/BE requests to improve memory utilization and reduce overheads. 
Extensive experiments validate the superiority of \sys{} over existing LLM serving systems~\cite{vllm, TGI}.

\bibliographystyle{plain}
\bibliography{references}


\end{document}